\documentclass[journal]{IEEEtran}

\usepackage{comment}
\usepackage{epsfig}
\usepackage{graphicx}
\usepackage{amsmath}
\usepackage{amssymb}
\usepackage{comment}
\usepackage{multirow}

\usepackage{hyperref}
\hypersetup{hidelinks}

\usepackage{booktabs}
\usepackage{array, caption, threeparttable}
\usepackage{caption}
\usepackage{subcaption}
\usepackage{algorithm}
\usepackage{listings}
\usepackage{color}
\usepackage{mathtools}
\usepackage{CJKutf8}

\usepackage{todonotes}
\usepackage{gensymb}

\begin{document}
\begin{CJK}{UTF8}{gbsn}

\title{rPPG-MAE: Self-supervised Pre-training with Masked Autoencoders for Remote Physiological Measurement}

\author{$\text{Xin Liu}^{\ast}$,~\IEEEmembership{Senior Member,~IEEE}, $\text{Yuting Zhang}^{\ast}$, $\text{Zitong Yu}^{\dagger}$, 
Hao Lu, Huanjing Yue, \\
$\text{Jingyu Yang}^{\dagger}$,~\IEEEmembership{Senior Member,~IEEE}
\thanks{Manuscript received June 4, 2023}
\thanks{$\ast$Equal contribution. $\dagger$Corresponding authors: Zitong Yu (email: yuzitong@gbu.edu.cn), and Jingyu Yang (email: yjy@tju.edu.cn).}


\thanks{X. Liu is with the School of Electrical and Information Engineering, Tianjin University, Tianjin 300072, China, and also with Computer Vision and Pattern Recognition Laboratory, School of Engineering Science, Lappeenranta-Lahti University of Technology LUT, Lappeenranta 53850, Finland.} 

\thanks{Y. Zhang, H. Yue, J. Yang are with the School of Electrical and Information Engineering, Tianjin University, Tianjin 300072, China}

\thanks{Z. Yu is with Great Bay University, Dongguan 523000, China.}

\thanks{H. Lu is with the Information Hub, the Hong Kong University of Science and Technology (Guangzhou), Guangzhou 511453, China.} 

}

\markboth{IEEE TRANSACTIONS ON MULTIMEDIA}%
{Shell \MakeLowercase{\textit{et al.}}: Bare Demo of IEEEtran.cls for IEEE Journals}

\maketitle

\begin{abstract}
Remote photoplethysmography (rPPG) is an important technique for perceiving human vital signs, which has received extensive attention. For a long time, researchers have focused on supervised methods that rely on large amounts of labeled data. These methods are limited by the requirement for large amounts of data and the difficulty of acquiring ground truth physiological signals. To address these issues, several self-supervised methods based on contrastive learning have been proposed. However, they focus on the contrastive learning between samples, which neglect the inherent self-similar prior in physiological signals and seem to have a limited ability to cope with noisy. In this paper, a linear self-supervised reconstruction task was designed for extracting the inherent self-similar prior in physiological signals. Besides, a specific noise-insensitive strategy was explored for reducing the interference of motion and illumination. The proposed framework in this paper, namely rPPG-MAE, demonstrates excellent performance even on the challenging VIPL-HR dataset. We also evaluate the proposed method on two public datasets, namely PURE and UBFC-rPPG. The results show that our method not only outperforms existing self-supervised methods but also exceeds the state-of-the-art (SOTA) supervised methods. One important observation is that the quality of the dataset seems more important than the size in self-supervised pre-training of rPPG.  The source code is released at \href{https://github.com/linuxsino/rPPG-MAE}{https://github.com/linuxsino/rPPG-MAE}. 
\end{abstract}


\IEEEpeerreviewmaketitle

\section{Introduction}

\IEEEPARstart{H}{eart} rate (HR), heart rate variation rate (HRV), and respiratory rate (RF) contain a lot of human vital information which are important health indicators. In the past, these physiological signals are typically measured by electrocardiogram (ECG) and photoplethysmography (PPG). These traditional methods require direct contact with the body which limited the real-time monitoring of a person's vital information in a sensor-free environment. Without attaching sensors, non-contact remote heart rate monitoring (rPPG) by analyzing skin color changes in videos of patients' faces has become a hot research topic. 

At the early stage, many methods \cite{PCA,GuhaBalakrishnan2013DetectingPF,DanielMcDuff2014ImprovementsIR,POS2014,CHROM2013} explored various handcrafted feature for rPPG. Recently, a number of end-to-end supervised models \cite{DeepPhys2018,DL1,yu2019remotePhysNetreal, yu2020autohr,huang2021spatioheartrate} by employing 2D/3D Convolutional Neural Networks (CNN) to extract rPPG features have been designed. Meanwhile, a few non-end-to-end fully-supervised approaches \cite{STMap2020,bvpnet} are developed, which capture rPPG signals from the spatio-temporal map (STMap). Such supervised methods require a huge number of labeled data, while in the field of rPPG, the collection of large-scale data with golden-standard ground-truth signals is costly and privacy sensitive. Therefore, a few self-supervised methods \cite{ContrastPhys,ICCV2021Theway,KimJinman2021SelfsupervisedRL, yang2022simper} have been proposed to deal with that limitation. Nevertheless, such existing unsupervised methods for rPPG tasks focus on the contrastive learning between samples, which neglect the inherent self-similar prior in physiological signals, might not be robust in challenging scenarios (e.g., serious head movement).  

Since rPPG signals are subtle and easily drowned out by noise (e.g. lighting, motion, camera noise \text{et al.}), it is difficult to extract periodic information from raw video data in the manner of original data structures. This is why many rPPG methods still construct the neural network input in a specific way rather than directly employing raw data, such as the spatio-temporal map (STMap)~\cite{STMap2020}. As STMap contains rich physiological information, it has been used successfully in supervised-learning methods \cite{bvpnet,STMap2020,Dual-GAN2021}. Therefore, it is promising to treat STMap as input and design a novel self-supervised method focusing on the inherent self-similar prior in physiological signals for accurate rPPG measurement. However, it is tough to obtain such self-similar prior in a varied-light/heavy-movement dataset like VIPL-HR~\cite{VIPL}, that's why there are few self-supervised methods conduct experiments on VIPL-HR. Just then, the successful handcrafted algorithms \cite{POS2014, CHROM2013} draw our attention, such algorithms added with manual priori knowledge to overcome the non-robustness in complex motion scenarios and achieve successes in traditional era. In the era of data-driven deep learning, we wonder whether traditional handcrafted algorithms still play a role. Thus, we are inspired to design a series of noise-insesnsitive STMap augmentations for the challenge of various scenarios, especially, we named the STMap combined with POS and CHROM algorithms as PC-STMap.

In recent years, self-supervised learning has become a craze in computer vision, and plenty of methods have been proposed, such as pretext-task-based self-supervised learning algorithms \cite{pretext1,pretext2,pretext3,pretext4} and contrastive-learning-based models \cite{2020MoCo,chen2020simclr}. Nowadays,
a form of more general denoising auto-encoders has achieved tremendous success in both natural language processing (NLP) (e.g., masked autoencoding for BERT~\cite{BERT}) and computer vision (e.g. masked autoencoders (MAE)~\cite{KaimingHe2021MaskedAA}). In particular, MAE has proved to be effective in image analysis tasks (e.g. image classification and object segmentation). rPPG is a typical computer vision task, and how to leverage masked autoencoder to reduce the information redundancy and noise of STMap to achieve efficient rPPG measurement has naturally become a research focus.

However, in early studies, mask autoencoders were only used to pretrain natural images like ImageNet dataset. Furthermore, there exists a big gap between rPPG and natural image processing tasks: \textbf{1)} The physical information that focused by rPPG is different. Other tasks focus more on appearance while rPPG focuses more on chroma changes. \textbf{2)} The difficulty of identifying valid information varies. Extracting rPPG signals from facial videos is relatively tough due to much-uncorrelated noise and subtle physiological signals. Thus, a novel rPPG loss function based on Pearson correlation coefficient was designed in reconstruction stage for the purpose of extracting the periodic prior in physiological signals.

Based on the discussions above, we carefully considered the advantage of leveraging MAE to pre-train ViT~\cite{vit} on light-invariant STMap augmentations and propose a new masked self-supervised method for rPPG measurement (rPPG-MAE). The contributions are summarized as follows:

\begin{table*}[t]\footnotesize
    \begin{center}
    \caption{Summary of the representative rPPG measurement methods in terms of traditional, supervised learning,
and self-supervised learning categories. SNR: signal-to-noise-ratio; PERC: PERCentage of total measurements “PERC” that the pulse-rate from the reference sensor and the rate from the method under test was in good agreement; AUC: Area Under Curve; ANOVA: Analysis of Variance; MAE: mean absolute HR error; Mean and Std:mean and standard deviation of the HR error; RMSE: the root mean squared HR error; MER: the mean of error rate percentage; r: Pearson’s correlation coefficients; MAPE: mean average percentage error; SD: standard deviation.}
    \label{tab:related work}
    \centering
    \resizebox{0.95\textwidth}{!}{
    \begin{tabular}{p{1.5cm} p{2.5cm}  p{1.5cm} p{4cm} p{3cm}  p{4cm} p{3cm}}
    
     \toprule
     \multicolumn{1}{c}{Type} & \multicolumn{1}{c}{Method}& \multicolumn{1}{c}{Venue} &  \multicolumn{1}{c}{Representation} &  \multicolumn{1}{c}{Backbone} & \multicolumn{1}{c}{Benchmark dataset} & \multicolumn{1}{c}{Metrics} \\
    
     \midrule
     \multirow{21}{*}{Traditional} &  \multicolumn{1}{c}{GREEN~\cite{GREEN2008}} &  \multicolumn{1}{c}{Opt. Express'08} & ROI-based averaged + band-pass filtered R.G.B signals & \multicolumn{1}{c}{-}  & \multicolumn{1}{c}{-} & SNR\\
     \cmidrule(lr){2-7}
     ~ &  \multicolumn{1}{c}{PCA~\cite{PCA}} &  \multicolumn{1}{c}{FedCSIS'11} & ROI selcted + different channels
combination(RGB) + FIR band-pass filtered RGB signals  & \multicolumn{1}{c}{-}  & Together 10 white volunteers, 2 women and 8 men, of different age (20 - 64 years), were examined & -\\
\cmidrule(lr){2-7}
     ~ &  \multicolumn{1}{c}{ICA~\cite{ICA2011}} &  \multicolumn{1}{c}{IEEE TBME'11} & ROI-based averaged, temporally detrended and normalized RGB signals  & \multicolumn{1}{c}{-}  & 12 participants of both genders (four females), different ages (18–31 years) and skin color& Mean, SD, RMSE, r \\
    \cmidrule(lr){2-7}
      ~ &  \multicolumn{1}{c}{CHROM~\cite{CHROM2013}} &  \multicolumn{1}{c}{IEEE TBME'13} & Color difference signals in which this specular reflection component is eliminated  & \multicolumn{1}{c}{-}  & Large population of 117 stationary subjects & SNR \\
      \cmidrule(lr){2-7}
      ~ &  \multicolumn{1}{c}{PBV~\cite{de2014improvedPBV}} &  \multicolumn{1}{c}{Physiol. Meas.'14} &  A linear combination of mean RGB skin-pixel values  & \multicolumn{1}{c}{-}  & Six videos recorded in a gym, large population of 117 stationary subjects & PERC, SNR \\
      \cmidrule(lr){2-7}
      ~ &  \multicolumn{1}{c}{2SR~\cite{wang20152SR}} &  \multicolumn{1}{c}{IEEE TBME'15} & A subspace of skin-pixels is constructed in RGB space +  a rotation of spatial subspaces between subsequent frames  & \multicolumn{1}{c}{-} & 54 RGB video sequences& SNR, r, AUC, ANOVA \\
      \cmidrule(lr){2-7}
       ~ &  \multicolumn{1}{c}{POS~\cite{POS2014}} &  \multicolumn{1}{c}{IEEE TBME'17} &  Using the
plane orthogonal to the skin-tone in the temporally normalized RGB space & \multicolumn{1}{c}{-}  & MAHNOB-HCI & SNR\\
      
      \midrule
    
     \multirow{21}{*}{Supervised} &  \multicolumn{1}{c}{HR-CNN~\cite{HR-CNN}} & \multicolumn{1}{c}{BMVC'18} & RGB facial videos  &Two-step convolutional neural network (CNN) & COHFACE, MAHNOB, PURE, ECG-Fitness & RMSE, MAE, r \\
     \cmidrule(lr){2-7}
      ~ &  \multicolumn{1}{c}{DeepPhys~\cite{DeepPhys2018}}  & \multicolumn{1}{c}{ECCV'18} & Motion representation with normalized frame difference and a current frame &  Two-branch CNN & MAHNOB-HCI, RGB video, Infrared Video & MAE, SNR \\
      \cmidrule(lr){2-7}
    ~ &  \multicolumn{1}{c}{SynRhythm~\cite{SynRhythm}} &  \multicolumn{1}{c}{ICPR'18} &  Spatial-temporal map  & ResNet-18 & MAHNOB-HCI, MMSE-HR & ME, SD, RMSE, MER \\
    \cmidrule(lr){2-7}
    ~ &  \multicolumn{1}{c}{EVM-CNN~\cite{qiu2018evm}}  & \multicolumn{1}{c}{IEEE TMM'18} & ROI extracted, feature image via spatial decomposition and temporal filtering  &  Shallow CNN & MMSE-HR & Mean, RMSE, SD, MAPE, r \\
      \cmidrule(lr){2-7}
    ~ &  \multicolumn{1}{c}{PhysNet~\cite{yu2019remotePhysNetreal}}  &  \multicolumn{1}{c}{BMVC'19} & RGB facial videos & 3DCNN & OBF, MAHNOB-HCI & SD, RMSE, r \\
    \cmidrule(lr){2-7}
    ~ &  \multicolumn{1}{c}{RhythmNet~\cite{STMap2020}}  & \multicolumn{1}{c}{IEEE TIP'20} & Spatial-temporal map in YUV space &  ResNet-18 + GRU & MAHNOB-HCI, MMSE-HR, VIPL-HR & MAE, RMSE, MER, r, Mean, Std \\
    \cmidrule(lr){2-7}
    ~ &  \multicolumn{1}{c}{MTTS-CAN~\cite{MTTS-CAN}}  & \multicolumn{1}{c}{NIPS'20} & Motion representation with normalized frame difference and a averaged frame &  Two-branch temporal shift CNN & MMSE-HR, AFRL & MAE, RMSE, r, SNR \\
    \cmidrule(lr){2-7}
    ~ &  \multicolumn{1}{c}{PulseGAN~\cite{PulseGAN}}  &  \multicolumn{1}{c}{IEEE JBHI'21} & CHROM signals & Conv1d & UBFC-rPPG, PURE & MAE, RMSE, MER, r\\
    \cmidrule(lr){2-7}
    ~ &  \multicolumn{1}{c}{Dual-GAN~\cite{Dual-GAN2021}} &  \multicolumn{1}{c}{CVPR'21} & Spatial-temporal map & Customized CNN & UBFC-rPPG,PURE, VIPL-HR & MAE, RMSE, MER, r \\
    \cmidrule(lr){2-7}
    ~ &  \multicolumn{1}{c}{EfficientPhys~\cite{efficientphys}} &  \multicolumn{1}{c}{WACV'23} & RGB facial videos & 2DCNN and Swin transformer & UBFC-rPPG,PURE, MMSE & MAE, MAPE, RMSE, r \\
    \cmidrule(lr){2-7}
    ~ &  \multicolumn{1}{c}{Physformer++~\cite{yu2023physformer++}} &  \multicolumn{1}{c}{IJCV'23} & RGB facial videos & Temporal difference video transformer &  VIPL-HR,MAHNOB-HCI, MMSE-HR, OBF & MAE, RMSE, SD, r \\  
     \midrule
     \multirow{12}{*}{Self-supervised} &\multicolumn{1}{c}{Gideon21~\cite{ICCV2021Theway}}  &  \multicolumn{1}{c}{ICCV'21} & RGB facial videos  & 3DCNN & PURE, COHFACE,  MR-NIRP-Car, UBFC & MAE, RMSE, r \\
     \cmidrule(lr){2-7}
     ~&  \multicolumn{1}{c}{Contrast-Phys~\cite{ContrastPhys}}  &  \multicolumn{1}{c}{ECCV'22} & RGB facial videos and NIR videos  & 3DCNN & UBFC-rPPG,PURE, OBF, MMSE-HR, MR-NIRP (NIR) & MAE, RMAE, r\\
      \cmidrule(lr){2-7}
     ~ &  \multicolumn{1}{c}{SLF-RPM~\cite{wang2022SLF-RPM}} &  \multicolumn{1}{c}{AAAI'22} & RGB facial videos & 3D ResNet-18 & MAHNOB-HCI,UBFC-rPPG, VIPL-HR-V2 &  MAE, RMSE, SD, r\\
     \cmidrule(lr){2-7}
     ~ &  \multicolumn{1}{c}{SIMPER~\cite{yang2022simper}} &  \multicolumn{1}{c}{ICLR'23} & RGB facial videos & A variant of TS-CAN & UBFC-rPPG, PURE  &  MAE, MAPE\\
     \cmidrule(lr){2-7}
      ~ &  \multicolumn{1}{c}{SiNC~\cite{speth2023SiNC}} &  \multicolumn{1}{c}{CVPR'23} & RGB facial videos & 3DCNN without temporal dilations & UBFC-rPPG,PURE, DDPM &  MAE, RMSE, r\\
      \cmidrule(lr){2-7}
      ~ &   \multicolumn{1}{c}{\textbf{rPPG-MAE (Ours)}}   &  \multicolumn{1}{c}{2023} & Spatio-temporal map with POS and CHROM algorithm  &  Vison transformer &  UBFC-rPPG,PURE, VIPL-HR & MAE, RMSE, r  \\

    \bottomrule
    \end{tabular}
    }
    \end{center}
    \vspace{-1.2em}
\end{table*}

\begin{itemize}
    \item We propose the rPPG-MAE, which leverages mask autoencoder (MAE) for self-supervised ViT pre-training to mine the self-similarity of physiological signals. To the best of our knowledge, it is the first time that a self-supervised method based on mask autoencoder is explored in the field of rPPG.
    
    \item We explore several kinds of noise-insensitive reconstruction targets for rPPG-MAE. A new form of STMap augmentation, referred to as PC-STMap, was proposed and demonstrated outstanding performance on VIPL-HR~\cite{VIPL}.
    
    \item We design a new rPPG loss function to constrain the MAE pre-training task. The proposed rPPG loss is more suitable for pre-training than the original pixel reconstruction loss adopted in vanilla MAE~\cite{KaimingHe2021MaskedAA}, which enables ViT to efficiently learn the periodic information of rPPG signal.
    
    \item We conduct experiments on several evaluation protocols (e.g. linear probing, transfer learning, semi-supervised learning, cross-dataset testing) and verify the powerful generalization ability of pre-trained ViT. Based on the well pretrained model, rPPG-MAE outperforms the state-of-the-art methods in multiple physiological tasks including HR, HRV, and RF estimation.

\end{itemize}

\begin{figure*}[t]
    \centering
    \includegraphics[scale=0.5]{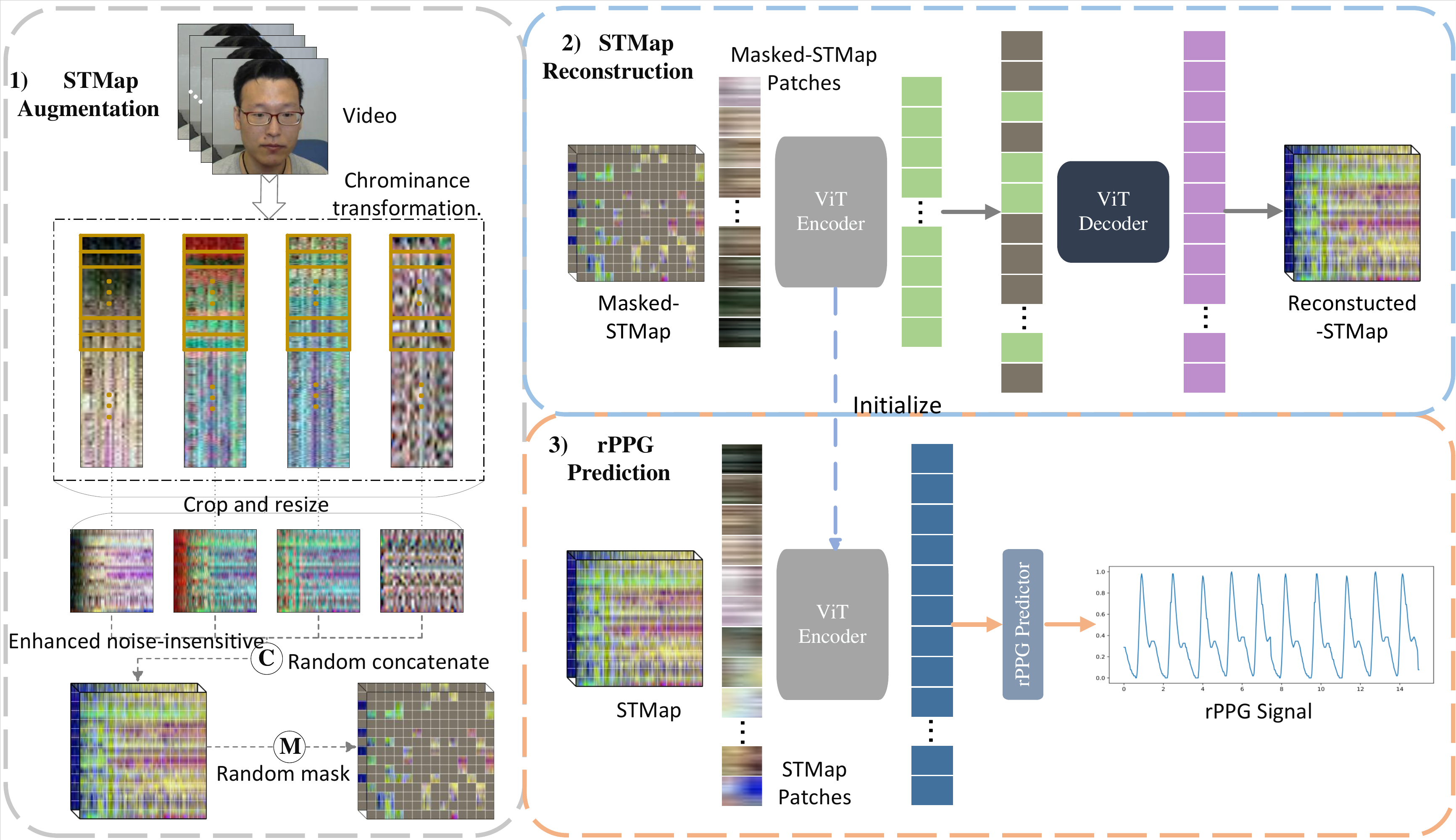}
      \caption{\small{Framework of the rPPG-MAE. It consists of STMap augmentation, STMap reconstruction and rPPG prediction. In STMap augmentation, the step of overlapping to resize the STMap is $s=5$, the size of STMap is $224\times224$ and the default mask ratio is $80\%$. In reconstruction, we pre-train the ViT encoder which learn the period information of rPPG signals. After that, the pre-trained ViT encoder is used in rPPG prediction. A linear layer (FC) is employed to predict rPPG signals.
      }
      }
    \label{fig:pipeline}
\end{figure*}

\section{Related Work}
\label{sec:relatedwork}

In this section, we have summarized some representative methods in recent years, as shown in Table \ref{tab:related work}. We first introduce some recent progress in remote physiological measurement, corresponding to the traditional and supervised section in Table \ref{tab:related work}. Then, previous self-supervised learning methods will be reviewed, corresponding to the self-supervised section in Table \ref{tab:related work}.

\subsection{Remote Physiological Measurement.}
\noindent\textbf{Traditional methods}\quad  Since 2008, Verkruysse \text{et al.} \cite{GREEN2008} first proposed the use of rPPG technology for physiological index measurement, a plenty of hand-crafted methods have been developed to extract stable BVP signals for rPPG measurement. They can be roughly divided into two categories: 1) Blind Source Separation (BSS)-based methods. They separate the most periodic source from the complex RGB temporal signals and treat it as a useful rPPG signal. Compared to \cite{GREEN2008}, such methods not only utilized the independent component analysis (ICA) \cite{ICA2011, lam2015robustICA} or principal component analysis (PCA) \cite{PCA, balakrishnan2013detectingPCA} to this field, but also proposed to select the ROIs from face areas for skin pixel values averaging in different channels rather than directly averaging the full face region. Additionally, the temporal filtering technique\cite{li2014remote, balakrishnan2013detectingPCA} has been demonstrated to be effective. 2) Projection-plane/subspace-based methods.These methods aim to find a specialized color space that contains most of the rPPG signals and suppress the noise outside this space. For example, CHROM \cite{CHROM2013} computes the pulse-signal as a linear combination of chrominance-signals assuming a standardized
skin-color to white-balance the camera. PBV \cite{de2014improvedPBV} defines a signature of blood volume change to distinguish the pulse-induced color changes from motion noise and compute another linear combination of mean RGB skin-pixel values. Before long, 2SR \cite{wang20152SR} estimate a spatial subspace of skin-pixels and measure its temporal rotation for pulse extraction. Another projection plane orthogonal to the skin tone was proposed in POS \cite{POS2014}. The main purpose of these methods is to improve the Signal-to-Noise-Ratio (SNR) of the pulse frequency. They do work to some extend. There are some limitations: BSS-baesd methods cannot deal with the case that motion is also periodic; the methods of the second category are based on some specific assumptions. They all have trouble dealing with complex situations.

\noindent\textbf{Supervised-learning methods}\quad With the great success of deep learning (DL) in various computer vision tasks~\cite{DL1}, supervised-learning methods have been studied for rPPG based physiological measurement~\cite{DL2}. In recent years, supervised-learning methods have mushroomed \cite{DL3,DL4,DL6,DL7,DL9,DeepPhys2018,Dual-GAN2021,HR-CNN,PhysNet,PulseGAN,STMap2020,PhysFormer,BMVC2018visual}. In terms of input, these methods can be divided into two folders, namely regular videos \cite{DeepPhys2018, MTTS-CAN,PhysNet,HR-CNN, yu2021facial} and STMap \cite{STMap2020,bvpnet,Dual-GAN2021}. As Transformer making great waves in CV, some researchers transferred Transfomer into rPPG feature representation learning. TransRPPG \cite{TransRPPG} extracted rPPG features from the preprocessed signal maps via ViT for face 3D mask presentation attack detection \cite{anti-spoofing}. Liu \text{et al.} created a pre-processing-free neural architecture named EfficintPhys-T, which is based on swin transformer\cite{swin}. Yu \text{et al.} \cite{PhysFormer} proposed the PhysFormer to capture long-range spatio-temporal attentional rPPG features from a facial video directly. After the release of \cite{PhysFormer}, an extended version called PhysFormer++ \cite{yu2023physformer++} was proposed shortly thereafter. Revanur \text{et al.} \cite{Instantaneous} proposed a video Transformer for estimating instantaneous heart rate and respiration rate from face videos. The above works validated Transformer has a strong ability to mine spatio-temporal attention from rPPG signal. However, the development of such supervised-learning approaches faces the dilemma that labels in the rPPG domain are difficult to obtain.

\subsection{Self-supervised learning.}
Self-supervised learning methods \cite{xu2020selfTMMVOS, KaimingHe2021MaskedAA} use data itself as supervisory information to learn the feature expression of sample data. Two common self-supervised learning methods are contrastive learning and mask autoencoder (MAE). There are several outstanding work \cite{2020MoCo,2021simsiam,grill2020byol,chen2020simclr,gao2021efficientTMMcontar, huo2021heterogeneousTMMcontra} in contrastive learning. These methods rely on the division of positive and negative sample pairs. MAE \cite{KaimingHe2021MaskedAA} introduced an asymmetric encoder and decoder architecture, where masked tokens are skipped in computation-heavy encoder and only pass all tokens through a light-weight decoder. MAE has proved to be effective in image classification and object segmentation. Later, researchers \cite{ibot,data2vec,maskfeat,peco} explored different reconstruction targets in MAE. For example, Data2Vec \cite{data2vec} predicts contextualized latent representations that contain information from the entire input instead of modality-specific targets. Maskfeat \cite{maskfeat} treats Histograms of Oriented Gradients (HOG) as target to pre-train model. These improvements are suitable for areas of image analysis such as image classification. 

Due to the advantages of self-supervised learning, many work in rPPG domain \cite{ContrastPhys,ICCV2021Theway,wang2022SLF-RPM, yang2022simper, speth2023SiNC} adopt self-supervised learning, such as the popular contrastive learning to train networks. \cite{ICCV2021Theway} uses contrastive learning with a weak prior over the frequency and temporal smoothness of the target signal of interest. To overcome the limitation of  extra computation burden and easily impacted by external periodic noise, \cite{ContrastPhys} proposed a new unsupervised method (Contrast-Phys) which bases on the rPPG similarity in both spatial and temporal domain. \cite{wang2022SLF-RPM} proposed two types of data augmentation to suit rPPG self-supervised learning more. \cite{yang2022simper} proposed a simple $\&$ effective SSL framework that learns periodic information in data. \cite{speth2023SiNC} applied the loss directly to the prediction by shaping the frequency spectrum, and encouraging variance over a batch of inputs. However, there are some limitations to these method: 1) they pay more attention to the relation or differences between samples, the characteristics of the sample itself are ignored. 2) they are vulnerable to noise (e.g. light change, motion) and fail to verify the effectiveness of the methods on the challenge dataset VIPL-HR. 

Different from the works mentioned above, the proposed rPPG-MAE is a self-supervised architecture specifically designed for the rPPG domain. We design a specific d light-invarient STMap augmentation and rPPG loss function for taking advantage of mask autoencoder, which demonstrate excellent performance even on the less-constrained VIPL-HR dataset.

\section{Methodology}
\label{sec:method}
In this section, we will introduce the rPPG-MAE for self-supervised pre-training on rPPG measurement. The main framework of rPPG-MAE is illustrated in Fig.~\ref{fig:pipeline}. First, the generation of STMap augmentations will be presented in Sec.~\ref{sec:Spatial-Temporal Map}. Then, we will introduce the STMap reconstruction task in Sec.~\ref{sec:Self-supervised Pretraining}, and finally the rPPG prediction task is discussed in Sec.~\ref{sec:rPPG Finetuning}.

\begin{figure*}
    \centering
    \includegraphics[scale=0.6]{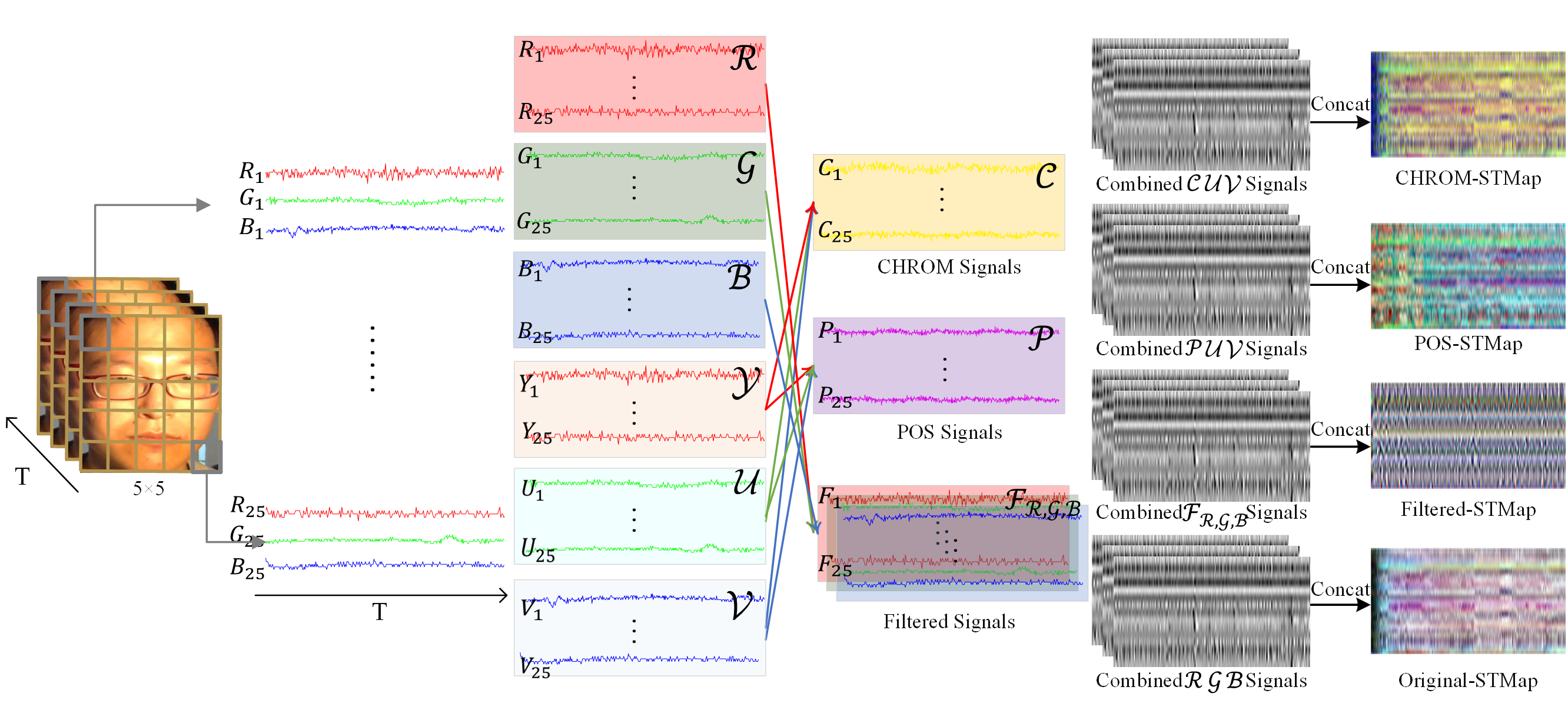}
      \caption{\small{
    Chrominance transformation. We first align faces across different frames based on the detected landmarks, then the facial area is divided
    into $n$ ROI blocks $R_1$,$R_2$,...$R_{25}$. The average color value is computed for each color channel within each block. The per-channel average color values of the same block location but different frames are concatenated into the sequences, i.e., $R_1$, $G_1$, $B_1$, $R_2$, $G_2$, $B_2$,...,$R_{25}$, $G_{25}$, $B_{25}$. The sequences from same color channel are concatenated into a map($\mathcal{R}$, $\mathcal{G}$, and $\mathcal{B}$) with size $N\times L$, where $N=25$. Besides, RGB color space transform to YUV color space($\mathcal{Y}$, $\mathcal{U}$, and $\mathcal{V}$) .Further, the CHROM signals and POS signals are processed using the CHROM~\cite{CHROM2013} algorithm and POS \cite{POS2014} algorithm. The filtered signals in the figure  ($\mathcal{F}_{\mathcal{R},\mathcal{G},\mathcal{B}}$) are filtered through a Butterworth bandpass filter with frequency of [0.6,3]. Finally, different combined signals are concatenated into four STMaps (CHROM-STMap, POS-STMap, Filtered-STMap, Original-STMap).
      } 
      }
    \label{fig:4maps}
\end{figure*} 

\subsection{Spatio-Temporal Map}
\label{sec:Spatial-Temporal Map}

Spatio-temporal map was proposed by Niu \text{et al.}~\cite{STMap2020}, which effectively combined the information of the physiological signal on human face. In brief, STMap is one kind of physiological map generated from several skin areas ROIs. We denote the $\mathit{i}$-th input STMap as $\mathit{\text{ST}^i}\in \mathbb{R}^{N\times T\times C}$, in which $\it{N}$ denotes the number of ROIs, $\it{T}$ denotes the frames number of a video clip, $\it{C}$ denotes the number of channels($C=3$, including R, G and B). As shown in Fig.~\ref{fig:pipeline}, we first generate a large STMap ($\text{ST}_L\in \mathbb{R}^{N\times T_L\times C}$) from a whole video, then crop the large STMap with a overlap of $\it{s}$ frames. The length of a clip is $\it{T}$. So, the number of STMap from a video is $N_{\text{st}}=(\it{T_L}-\it{T})/\it{s}+1$. After that, we resize the original STMap to $\text{ST}_r^i\in \mathbb{R}^{T\times T\times C}$, the number of ROIs increases from $N$ to $T$. 

\textbf{Noise-insensitive STMap augmentations.} To mitigate the effects of light and ambient noise, on the basis of original STMap, we leverage POS~\cite{POS2014}, CHROM~\cite{CHROM2013} algorithm and Butterworth band-pass filter for several types of STMap augmentations. As shown in Fig. \ref{fig:4maps}, POS-STMap~\cite{POS-STMap} consists of U and V channels from the facial videos and a POS channel processed using POS algorithm \cite{POS2014}. CHROM-STMap is similar to POS-STMap, the only difference is that the POS channel is replaced by CHROM channel, which processed by CHROM algorithm \cite{CHROM2013}. For Filterd-STMap, we bandpass each channel of STMap and each ROI with a frequency range of [0.6, 3] Butterworth filter. The four types of STMap can be expressed as
\begin{equation}
    \begin{split}
    \text{CHROM-STMap}&=\mathbf{C}(\tiny{\mathcal{C}_{(\mathcal{R},\mathcal{G},\mathcal{B})},\mathcal{R}, \mathcal{G}}),\\
    \text{POS-STMap}&=\mathbf{C}(\tiny{\mathcal{P}_{(\mathcal{R},\mathcal{G},\mathcal{B})},\mathcal{R},\mathcal{G}}),\\
    \text{Filtered-STMap}&=\mathbf{C}(\tiny{\mathcal{F}_{\mathcal{R}},\mathcal{F}_{\mathcal{G}},\mathcal{F}_{\mathcal{B}}}),\\
    \text{Original-STMap}&=\mathbf{C}(\tiny{\mathcal{B},\mathcal{R},\mathcal{G}}),
    \end{split}
\end{equation}

where $\mathcal{C}$, $\mathcal{P}$, $\mathcal{F}$ indicate the CHROM algorithm，POS algorithm, and Butterworth filter, respectively; $\mathcal{R}$, $\mathcal{G}$, $\mathcal{B}$ denote the three color channels of signals from different ROIs; $\mathbf{C}$ (·) indicates the operation of concatenating.

\begin{figure}
\centering
\includegraphics[scale=0.5]{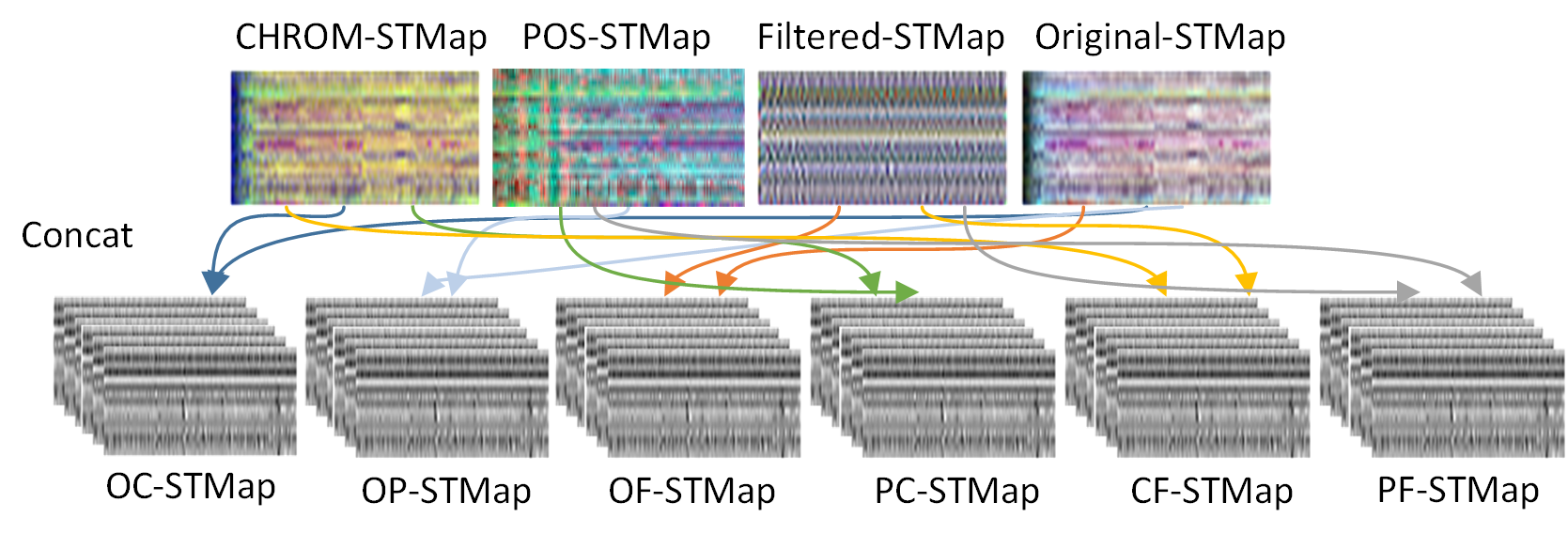}
  \caption{\small{The generation of six enhanced noise-insensitive STMaps. The enhanced version combines various physical prior to resist the light interference and motion noise.
  }
  }
\label{fig:figmaptype}
\vspace{-0.8em}
\end{figure}

Based on this, we explore six enhanced STMap augmentations, with the hope of combining the advantage of several chromas and algorithms to achieve stronger robustness. The generation of the six enhanced variants of STMap is shown in  Fig. \ref{fig:figmaptype}, which can be represented as:
\begin{equation}
    \begin{split}
    \text{OC-STMap}&=\mathbf{C}(\tiny{\text{Original-STMap}, \text{CHROM-STMap}}),\\
    \text{OP-STMap}&=\mathbf{C}(\tiny{\text{Original-STMap}, \text{POS-STMap}}),\\
    \text{OF-STMap}&=\mathbf{C}(\tiny{\text{Original-STMap}, \text{Filtered-STMap}}),\\
    \text{PC-STMap}&=\mathbf{C}(\tiny{\text{POS-STMap}, \text{CHROM-STMap}}),\\
    \text{CF-STMap}&=\mathbf{C}(\tiny{\text{CHROM-STMap}, \text{Filtered-STMap}}),\\
    \text{PF-STMap}&=\mathbf{C}(\tiny{\text{POS-STMap}, \text{Filtered-STMap}}),\\
    \vspace{-0.3em}
    \end{split}
\end{equation}
where Original-STMap, CHROM-STMap, POS-STMap, Filtered-STMap and $\mathbf{C}$ are mentioned above.

\textbf{Masked-STMap.} In reconstruction stage, the masked-STMaps are needed as the input. Since the periodic signal is uniformly distributed in the time domain, we choose to random mask STMap. The implementation strategy of masking is straightforward: we first divide the STMap into non-overlapping patches, then shuffle the patches, at last orderly keep the patches in proportion and then remove the remaining patches. The reserved patches are the input in reconstruction stage for self-supervised pre-training. The number of reserved patches can be calculated via $L_{k}=(1-R_{m})\cdot(T/P_s)^2$, where $R_{m}$, $T$, $P_s$ indicate the mask ratio, the length of input clip, and the patch size of ViT encoder, respectively.

\subsection{STMap Reconstruction}
\label{sec:Self-supervised Pretraining}

\textbf{ViT encoder.} Following model of ViT~\cite{vit}, our ViT encoder embeds patches by a linear projection with added positional embeddings, and then processes the STMap patches via several Transformer blocks. In the proposed method, we choose the base version of ViT which has 12 Transformer blocks, and 768 dimensions. The input of ViT encoder in the self-supervised pre-training stage are the reserved patches ($P_k^i$) in masked-STMap which we mentioned above. The ViT encoder output $X_{\text{ViE}}^i=\mathbf{E}_{\text{ViE}}(P_k^i)$, where $X_{\text{ViE}}^i\in \mathbb{R}^{L_{k}\times D_e}$, $L_k$ and $D_e$ indicate the length of input STMap sequence and the dimension of the ViT encoder.

\textbf{ViT decoder.} The output of ViT encoder is just an encoding of the reserved patches. In order to reconstruct STMap, a ViT decoder is required. The numbers of blocks and dims are two points needed to be discussed, the details are given in Sec. \ref{sec:Ablation Studies}. But it is certain that the ViT decoder is smaller than the ViT encoder. In the proposed model, the default ViT decoder has 8 Transformer blocks and 128 dims. Different from ViT encoder, the inputs of ViT decoder are the whole patches consisting of encoded patches and mask tokens. The mask tokens are some shared and learnable vectors with positional embeddings, after ViT decoder, they will be predicted as the missing patches of STMap. Thus, the length of output is the number of the patches in a whole STMap ($L_{\text{all}}=(T/P_s)^2$. The ViT decoder output $X_{\text{ViD}}^i=\mathbf{E}_{\text{ViD}}(X_{\text{ViE}}^i)$, where $X_{\text{ViD}}^i\in \mathbb{R}^{L_{\text{all}}\times D_d}$, $L_{\text{all}}$ and $D_d$ indicate the length of the whole STMap sequence and the dimension of the ViT decoder, respectively. However, the default dim $D_d$ is not match to the number of pixel values in a patch, so that a linear projection is designed at the last layer of decoder. In that way, a token is reshaped into a patch, and then we can obtain the desired reconstructed STMap.

\textbf{Reconstruction loss.} The reconstruction task aims to ensure that ViT encoder learns the periodic characteristics of BVP signals by reconstructing a new STMap. The output of ViT decoder is a series of vectors whose dimension is equal to the number of pixels in a patch. The pixel loss function only computes the mean squared error (MSE) between the reconstructed and original images in the masked pixel space, as
\vspace{-0.3em}
\begin{equation}
    \mathcal{L}_{\text{pixel}}=\text{MSE}(P_{m_{\text{pr}}}, P_{m_{\text{gt}}}),
    \vspace{-0.3em}
\end{equation}
where $P_{m_{pr}}$ means the masked pixel values predicted by ViT decoder, $P_{m_{gt}}$ means the ground-truth masked pixel values of STMap, and $MSE(\cdot)$ denotes the mean squared error.

To enable the ViT encoder to learn the periodicity of BVP signal, we propose a new loss function
\vspace{-0.3em}
\begin{equation}
    \mathcal{L}_{\text{rPPG}}= \frac{\sum_{i=1}^C \sum_{j=1}^{N_{\text{ROI}}} (1-\text{PC}(P_{\text{pr}}^{i,j}, P_{\text{gt}}^{i,j}))}{C\times N_{\text{ROI}}},
    \vspace{-0.3em}
\end{equation}
where $P_{\text{pr}}^{i,j}$ and $P_{\text{gt}}^{i,j}$ indicate the pixel values of a row of the reconstructive STMap and ground-truth STMap, respectively. The $\text{PC}(\cdot)$ denotes the Pearson correlation. $C$ and $N_{\text{ROI}}$ represent the number of channels and ROI, respectively, where $N_{\text{ROI}}=T$.

\begin{figure}
\centering
\includegraphics[scale=0.35]{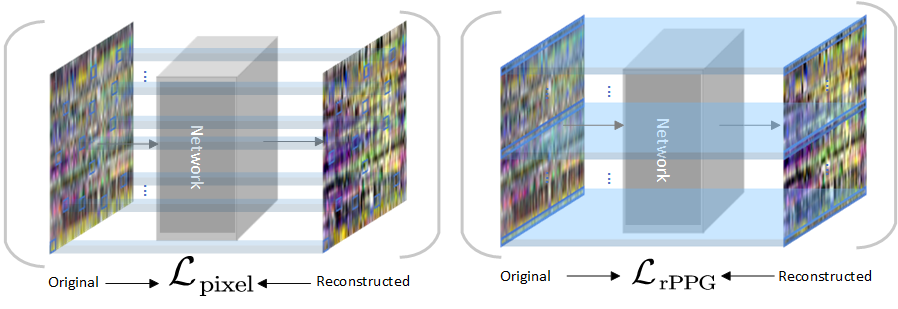}
  \caption{\small{Visualization of reconstructive loss.
  }
  }
\label{fig:loss}
\vspace{-0.4em}
\end{figure}

In summary, the overall loss function for self-supervised pre-training stage can be written as
\vspace{-0.3em}
\begin{equation}
    \mathcal{L}_{\text{overall}}= \lambda \mathcal{L}_{\text{pixel}} + (1-\lambda)\mathcal{L}_{\text{rPPG}},
    \vspace{-0.3em}
\end{equation}
where the hyper-parameter $\lambda\in \{0,1\}$. Fig. \ref{fig:loss} shows the visualization of the two reconstructive loss. Pixel loss is the regression of discrete pixel points, while rPPG loss ensures the continuity of signal in the time dimension.

\begin{table*}[t]\footnotesize
    \begin{center}
    \caption{Intra-dataset HR results. The best results are in bold, and the second-best
    results are underlined. MeanAE: Mean Absolute Error; RMSE: Root Mean Square Error; r: Pearson correlation coefficient. $\star$ Methods in the self-supervised row require fine-tuning on labeled data with a linear classifier. Baseline$\blacktriangle$ denotes the results of fine-tuning on three datasets with ImageNet pre-trained weights initialized ViT encoder, respectively.  }
    \label{tab:Intra HR}
    \centering
    \resizebox{1\textwidth}{!}{\begin{tabular}{p{1.5cm} p{2.5cm}  p{0.5cm} p{0.5cm} p{0.6cm}  p{0.5cm} p{0.5cm} p{0.6cm} p{0.5cm} p{0.5cm} p{0.6cm} }
    
     \toprule
      \multirow{2}{*}{Type}& \multirow{2}{*}{Method}& \multicolumn{3}{c}{VIPL-HR} &  \multicolumn{3}{c}{PURE} &  \multicolumn{3}{c}{UBFC-rPPG} \\
      
    \cmidrule(lr){3-5} \cmidrule(lr){6-8} \cmidrule(lr){9-11} 
    
    ~& ~&\multicolumn{1}{c}{MeanAE $\downarrow$} & \multicolumn{1}{c}{RMSE $\downarrow$} & \multicolumn{1}{c}{$r$ $\uparrow$}  &  \multicolumn{1}{c}{MeanAE $\downarrow$} & \multicolumn{1}{c}{RMSE $\downarrow$} & \multicolumn{1}{c}{$r$ $\uparrow$}  & \multicolumn{1}{c}{MeanAE $\downarrow$} &\multicolumn{1}{c}{RMSE $\downarrow$} & \multicolumn{1}{c}{$r$ $\uparrow$} \\
     \midrule
     \multirow{4}{*}{Traditional} & GREEN~\cite{GREEN2008} & -& -& -  & - & - & -  & 7.50& 14.41 & 0.62 \\
     ~ & ICA~\cite{ICA2011} & - & -  & -  & -  & - & - & 5.17  & 11.76 & 0.65 \\
     ~ & POS~\cite{POS2014} & 11.5 & 17.2  & 0.30  & -  & - & - & 4.05  & 8.75 & 0.78 \\
      ~ & CHROM~\cite{CHROM2013} & 11.4 & 16.9  & 0.28  & 2.07  & 9.92 & \underline{0.99} & 2.37  & 4.91 & 0.89 \\
      \midrule
    
     \multirow{5}{*}{Supervised} & HR-CNN~\cite{HR-CNN} &- &-  &- & 1.84  & 2.37 & 0.98 & -  & - & -\\
    ~ & SynRhythm~\cite{SynRhythm} &- & - & - &- & -& -& 5.59& 6.82& 0.72\\
    ~ & PhysNet~\cite{PhysNet}  &10.8 & 14.8 & 0.2 &2.1 & 2.6& \underline{0.99}& -& -& -\\
    ~ & PulseGAN~\cite{PulseGAN}  &- & - & - &- & -& -& 1.19& 2.10& \underline{0.98}\\
    ~ & RhythmNet~\cite{STMap2020}  &5.3 & 8.14 & \underline{0.76} &- & -& -& 1.19& 2.10& \underline{0.98}\\
    ~ & Dual-GAN~\cite{Dual-GAN2021} &\underline{4.93} & \underline{7.68} & \textbf{0.81} &0.82 & \underline{1.31} & \underline{0.99} & 0.44& 0.67 & \textbf{0.99}\\
    ~ & \textbf{Baseline$\blacktriangle$} &6.71 & 10.37 & 0.58
    &18.79 & 25.08 & 0.31 & \underline{0.24}& \underline{0.29} & \textbf{0.99}\\
     \midrule
    
     \multirow{6}{*}{Self-supervised} &  Gideon21~\cite{ICCV2021Theway}  & 9.80 & 15.48  & 0.38 & 2.1  & 2.6 & \underline{0.99} & 3.6  & 4.6 & 0.95 \\
     ~& Contrast-Phys~\cite{ContrastPhys}  & 7.49 & 14.40  & 0.49 & 1.00  & 1.40 & \underline{0.99} & 0.64  & 1.00 & \textbf{0.99}\\
     ~ & SimPer~\cite{yang2022simper}$\star$ &-&-&-&3.89&-&-&4.24&-&-\\
     ~& SLF-RPM~\cite{wang2022SLF-RPM}$\star$ &-&-&- &-&-&- &8.39 & 9.70 & 0.70\\
     ~& SiNC~\cite{speth2023SiNC} &-&-&- &\underline{0.61} &1.84 &\textbf{1.00} & 0.59 & 1.83 & \textbf{0.99}\\

      ~ &  \textbf{rPPG-MAE (Ours)}$\star$   & \textbf{4.52} & \textbf{7.49}  & \textbf{0.81} & \textbf{0.40}  & \textbf{0.92} & \underline{0.99} &  \textbf{0.17}  & \textbf{0.21} & \textbf{0.99} \\
    \bottomrule
    \end{tabular}}
    \end{center}
\end{table*}

\begin{table*}[t]\footnotesize
    \begin{center}
    \caption{HRV results on UBFC-rPPG. The best results are in bold, and the second-best results are underlined. RMSE: Root Mean Square Error; r: Pearson correlation coefficient. STD: standard deviation Baseline$\blacktriangle$ denotes the HRV results of fine-tuning on UBFC-rPPG dataset with ImageNet pre-trained weights initialized ViT encoder.  }
    \label{tab:HRV}
    \centering
    \resizebox{1\textwidth}{!}{\begin{tabular}{p{1.8cm} p{2.5cm}  p{0.5cm} p{0.5cm} p{0.6cm}  p{0.5cm} p{0.5cm} p{0.6cm} p{0.5cm} p{0.5cm} p{0.6cm} p{0.5cm} p{0.5cm} p{0.6cm}}
    
    \toprule
     \multirow{2}{*}{Type} & \multirow{2}{*}{Method}
     &  \multicolumn{3}{c}{LF (n.u.)} &  \multicolumn{3}{c}{HF (u.n)} &  \multicolumn{3}{c}{LF /HF} &  \multicolumn{3}{c}{RF (HZ)}\\
      
    \cmidrule(lr){3-5} \cmidrule(lr){6-8}
    \cmidrule(lr){9-11} \cmidrule(lr){12-14}

     \multicolumn{1}{c}{~} & \multicolumn{1}{c}{~}   &  \multicolumn{1}{c}{STD} & RMSE & \multicolumn{1}{c}{$r$}  &  \multicolumn{1}{c}{STD} & RMSE & \multicolumn{1}{c}{$r$}  & \multicolumn{1}{c}{STD} & RMSE & \multicolumn{1}{c}{$r$}  &  \multicolumn{1}{c}{STD} & RMSE & \multicolumn{1}{c}{$r$}\\
     \midrule
     \multirow{3}{*}{Traditional} & GREEN~\cite{GREEN2008} & 0.186 & 0.186  & 0.280  & 0.186  & 0.186 & 0.280 & 0.361  & 0.365 & 0.492 & 0.087  & 0.086 & 0.111\\
     
     ~ & ICA~\cite{ICA2011} &0.243	&0.240	&0.159	&0.243	&0.240	&0.159	&0.655	&0.645	&0.226	&0.086&	0.089 &0.102\\
     
     ~ & POS~\cite{POS2014} &0.171	&0.169	&0.479	&0.171	&0.169	&0.479	&0.405	&0.399	&0.518	&0.109	&0.107	&0.087\\
      \midrule
    
    \multirow{2}{*}{Supervised} & CVD~\cite{CVD}    &0.053 & 0.056 & 0.740 & 0.053 & 0.065 & 0.740 & 0.169 & 0.168 & 0.812 & 0.017 & 0.018 & 0.252\\
    ~& Dual-GAN ~\cite{Dual-GAN2021} &\textbf{0.034} & \textbf{0.035} & \textbf{0.891} & \textbf{0.034} & \textbf{0.034} & \textbf{0.891} & \underline{0.131} & \underline{0.136} & \underline{0.881}& \underline{0.010} & \underline{0.010} & \underline{0.395}\\
    ~& \textbf{Baseline$\blacktriangle$} &0.061 &0.062 &0.318 &0.061 &0.062 &0.318 &0.198 &0.185 &0.441 &0.008 &0.008 &0.443\\
    
     \midrule
    
     \multirow{4}{*}{Self-supervised} & Gideon21~\cite{ICCV2021Theway}   &0.091 &0.139 &0.694 &0.091 &0.139 &0.694 &0.525 &0.691 &0.684 &0.061 &0.098 &0.103\\
     
     ~& Contras-Phys~\cite{ContrastPhys}   &0.050 &0.098 &0.798 &0.050 &0.098 &0.798 &0.205 &0.395 &0.782 &0.055 &0.083 &0.347\\
     ~ &  \textbf{rPPG-MAE (Ours)}  & \underline{0.036} & \underline{0.037} & \underline{0.826} & \underline{0.036} & \underline{0.037} & \underline{0.826} & \textbf{0.124} & \textbf{0.130} & \textbf{0.887} & \textbf{0.007} & \textbf{0.007} & \textbf{0.522}\\
     
    \bottomrule
    \end{tabular}}
    \end{center}
\end{table*}

\subsection{rPPG Prediction}
\label{sec:rPPG Finetuning}

The rPPG prediction stage consists of a ViT encoder pretrained in the reconstruction task, and a rPPG predictor, in which a linear layer (as the rPPG predictor) used to predict rPPG signal. During rPPG prediction stage, the pretrained ViT encoder is fine-tuned, and the rPPG predictor is trained. Different from the reconstruction stage, the input to this stage is the full patches of STMap ($P_{\text{all}}^i$). Given the ViT encoder output in this stage $X_{\text{ViE}_r}^i=\mathbf{E}_{\text{ViE}_r}(P_{\text{all}}^i)$, where $X_{\text{ViE}_r}^i\in \mathbb{R}^{L_{\text{all}}\times D_e}$, and $L_{\text{all}}$ and $D_e$ indicate the length of the whole STMap sequences and the dimension of the ViT encoder, respectively. To predict the rPPG signal, a Negative Pearson correlation loss calculated between the predicted rPPG signal and the ground-truth BVP signal is selected, which can be denoted as
\vspace{-0.3em}
\begin{equation}
    \mathcal{L}_p=1-\text{PearC}(S_{\text{pr}},S_{\text{gt}}),
\vspace{-0.3em}
\end{equation}
where $S_{\text{pr}}$ and $S_{\text{gt}}$ denote the predicted rPPG signal and the ground-truth BVP signal, respectively. PearC($\cdot$) indicates Negative Pearson correlation.

In addition, a frequency domain loss is utilized for a better prediction, which compute the cross-entropy error between the ground-truth HR and the spectral distribution of the estimated rPPG signal, denoted as
\vspace{-0.3em}
\begin{equation}
    \mathcal{L}_{\text{fre}}=\text{CE}(\text{PSD}(s_{\text{pr}}),O_{\text{gt}}),
\vspace{-0.3em}
\end{equation}
where PSD(·) denotes the power spectral density of the predicted rPPG signal($s_{pr}$), and CE(·) denotes the cross-entropy loss. The ground-truth HR can be represented by a one-hot vector HR = [0, ..., 0, 1, 0, ...], and ‘1’ denotes the index corresponding to ground-truth HR. 

In a whole, the overall loss function for rPPG prediction stage can be written as :
\vspace{-0.3em}
\begin{equation}
    \mathcal{L}_{\text{all}}=\gamma \mathcal{L}_p + (1-\gamma)\cdot(\mathcal{L}_{\text{fre}}),
\vspace{-0.3em}
\end{equation}
where parameter $\gamma\in \{0,1\}$, which will be adjusted between different datasets. In our experiments, we set $\gamma=0$ in VIPL-HR~\cite{VIPL} dataset, $\gamma=1$ in both PURE~\cite{PURE} and UBFC-rPPG~\cite{UBFC} datasets.

\section{Experiments}
\label{sec:experiment}

In the experiments, three types of physiological signals for rPPG-based physiological measurement, i.e., heart rate (HR), heart rate variability (HRV), and respiration frequency (RF), are evaluated on three public-domain datasets (VIPL-HR~\cite{VIPL}, PURE~\cite{PURE}, and UBFC-rPPG~\cite{UBFC}).

\subsection{Datasets and Performance Metrics}
\label{sec:dataset}

\textbf{VIPL-HR} is a large-scale dataset which contains 2,378 visible light videos (VIS) and 752 near-infrared (NIR) videos of 107 subjects. The frame rate of the video varies depending on the device. \textbf{PURE} contains 60 RGB videos from 10 subjects that were recorded in 6 different setups. \textbf{UBFC-rPPG} comprises two datasets that are focused specifically on rPPG analysis. The first dataset is composed of 8 videos (about 16,500 frames), and the second one contains 42 RGB videos recorded with a Logitech C920 HD Pro webcam in a resolution of $640\times480$ with 30 FPS. We normalize the videos and the corresponding BVP signals of all three datasets to 30 HZ by cubic spline interpolation.

\textbf{Metrics.} We perform average HR estimation on all three datasets. Specifically, we follow existing methods~\cite{Dual-GAN2021,PhysFormer} and report low frequency (LF), high frequency (HF), and LF/HF ratio for HRV and RF estimation on high-quality UBFC-rPPG dataset. The most commonly used performance metrics for evaluation, including the standard deviation (STD), mean absolute error (MAE), root mean square error (RMSE), and Pearson's correlation coefficient ($r$) are reported. 
\subsection{Implementation Details}
\label{sec:Details}
The proposed method is implemented with Pytorch. We pretrain the ViT encoder in reconstruction task for 400 epochs, and finetune on all three datasets for 20 epochs separately, and only linear probe on VIPL-HR dataset for 50 epochs. 

\textbf{Pre-processing.} Following \cite{STMap2020}, we use an open source face detector SeetaFace to detect face and localize 81 facial landmarks, and then a face bounding box is designed to align face area and remove background area.

\textbf{Pre-training.} We divide each of three datasets into 5 folds, 4 folds of which is training set and the remaining 1 fold is testing set. We pretrain ViT by feeding with training set, and the AdamW optimizer is used with default betas of 0.9 and 0.95, weight decay of 0.05, learning rate schedule of cosine decay, warmup epochs of 40, and base learning rate of 0.001. We use the linear lr scaling rule $lr = base lr\times batchsize / 256$. The batch size is 64.

\textbf{Fine-tuning.} We finetune the weights of ViT encoder and rPPG predictor with labels. The default settings are as follows, AdamW optimizer with default betas of 0.9 and 0.9999, layer decay of 0.75, weight decay of 0.05, learning rate schedule of cosine decay, warmup epochs of 5 and base learning rate of 0.001. The learning rate scaling rule and batch size are the same to pretraining stage. 

\textbf{Semi-supervised learning.} we keep the training set and testing set the same with pre-training protocal. Different from full self-supervised, we divided the training set with several persentages (e.g. 10$\%$, 20$\%$, 50$\%$) in our semi-supervised experiments. Take 10$\%$ for example, the 90$\%$ of the training set are used for pre-training (without label) and the rest 10$\%$ of the training set are used for finetuning (with label). The configuration of the hyperparameter is the same with finetuning.

\textbf{Linear probing.} To estimate the effectiveness of the ViT encoder, we freeze the weights of self-supervised ViT encoder layers, and use class token instead of global pool for classification. The default settings are slightly different from fine-tuning, i.e., AdamW optimizer with default betas of 0.9, no weight decay, and base learning rate of 0.01, batch size is 512.

\begin{figure}
    \centering
    \includegraphics[scale=0.5]{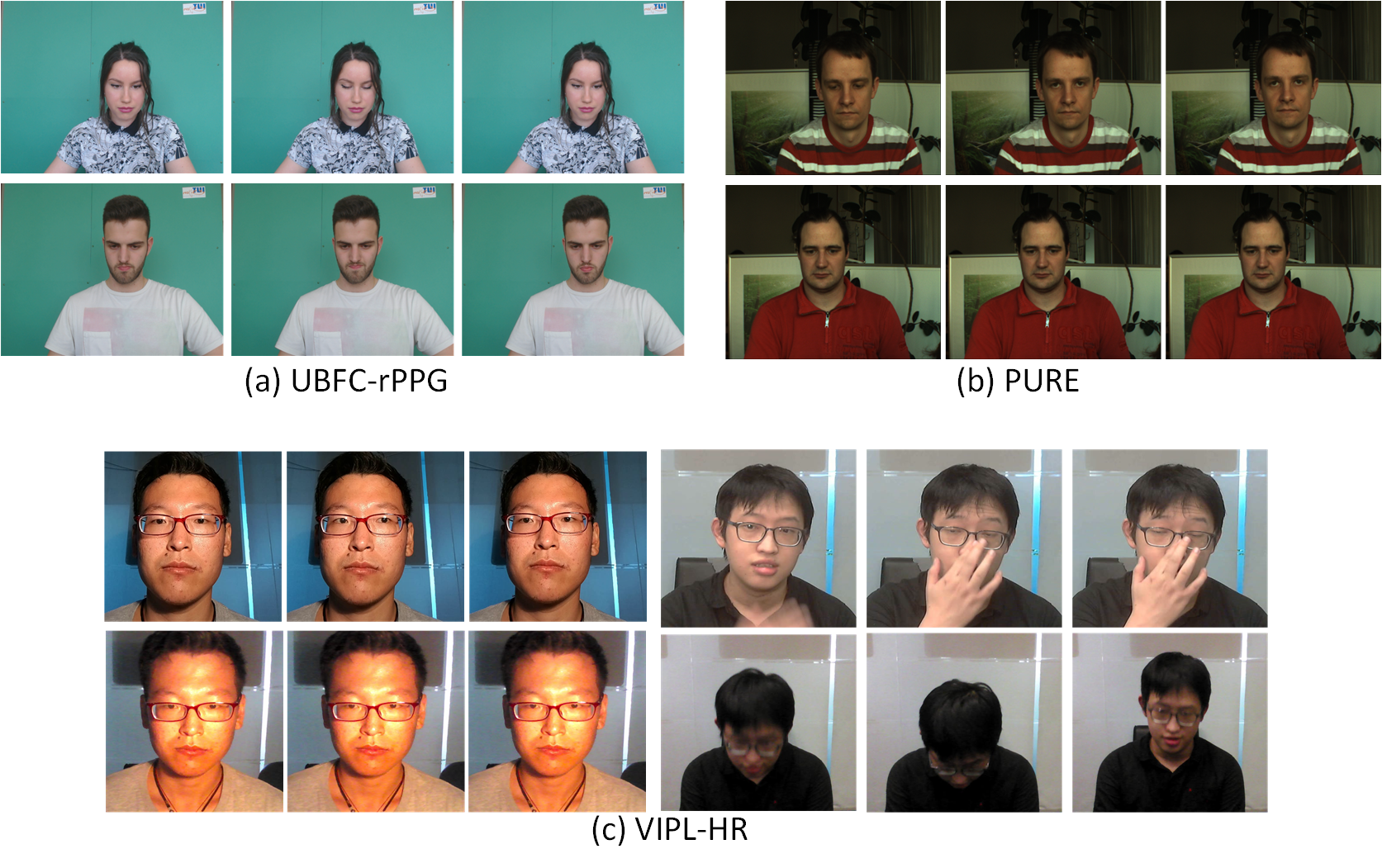}
      \caption{\small{Example video frames of different dataset.
      }
      }
    \label{fig:dataset}
\end{figure} 

\subsection{Intra-dataset Testing}
\label{sec: intra-database}
\textbf{HR estimation on three benchmark datasets.} The effectiveness of the pretrained ViT encoder is evaluated via performing HR estimation on VIPL-HR, PURE, and UBFC-rPPG datasets. As shown in Table \ref{tab:Intra HR}, we compare the proposed method with four traditional methods (GREEN~\cite{GREEN2008}, ICA~\cite{ICA2011}, POS~\cite{POS2014}, and CHROM~\cite{CHROM2013}); seven supervised DL methods (HR-CNN~\cite{HR-CNN}, SynRhythm~\cite{SynRhythm}, PhysNet~\cite{PhysNet}, PulseGAN~\cite{PulseGAN}, RhythmNet~\cite{STMap2020}, Dual-GAN~\cite{Dual-GAN2021} and Physformer++~\cite{yu2023physformer++}); five self-supervised methods (Gideon21~\cite{ICCV2021Theway}, Contrast-Phys~\cite{ContrastPhys}, SIMPER~\cite{yang2022simper}, SLF-RPM~\cite{wang2022SLF-RPM}, and SiNC~\cite{speth2023SiNC}). Except for the results of Contrast-Phys and Gideon21 on VIPL-HR dataset that we implement ourselves, all the other results are come directly from \cite{ICCV2021Theway} and \cite{ContrastPhys} or the corresponding original papers. From Table \ref{tab:Intra HR}, we can see that the proposed method outperforms all other methods on PURE and UBFC-rPPG and VIPL-HR. It should be noted that the results on VIPL-HR are based on PC-STMap and the results on the other datasets are based on the original STMap. Since PURE and UBFC-rPPG are datasets with simple scenarios, the specialized processing is not needed.  All of our proposed models show excellent experiments on UBFC-rPPG, due to the fact that the dataset itself of high quality. However, on a more complex PURE dataset, using STMap pretrained ViT encoder in a self-supervised manner seems to work, which makes the MAE score jumped from 18.79 to the highest value of 0.40. On the largest VIPL-HR dataset, the proposed method showed a better result even compared to the full supervised methods. From the experimental results, it is worth pointing out that the ViT encoder pre-trained with STMap is significantly better than that pre-trained with ImageNet, which verify the effectiveness of our self-supervised pretraining.

\begin{table}[t]\large
    \centering
    \caption{Transfer learning on three banchmark datasets.} 
    \label{tab:TransferL}
    
    
    \resizebox{0.45\textwidth}{!} {\begin{tabular}{p{3cm}  p{1.7cm} p{1.5cm} c c c  } 
     \toprule
     \multirow{2}{*}{Srategy} &\multirow{2}{*}{Pretraining}&\multirow{2}{*}{Fine-tuning}&  \multicolumn{3}{c}{HR (bpm)}\\
     \cmidrule(lr){4-6}
     ~&~&~& \multicolumn{1}{c}{MeanAE $\downarrow$} & \multicolumn{1}{c}{RMSE $\downarrow$} & \multicolumn{1}{c}{$r$ $\uparrow$} \\
     \midrule
     \multirow{10}{*}{Transfer Learning}&\multirow{3}{*}{ImageNet}& \multicolumn{1}{c}{VIPL} & 6.71 & 10.37 & 0.58  \\ 
     ~&~ &\multicolumn{1}{c}{PURE}  & 18.79& 25.08 & 0.31 \\
     ~&~ & \multicolumn{1}{c}{UBFC} & 0.24 & 0.29 & 0.99 \\
     \cmidrule(lr){2-6}
     
     ~&\multirow{3}{*}{VIPL} & \multicolumn{1}{c}{VIPL}  & 5.57 & 9.10 & 0.70\\
     ~&~ & \multicolumn{1}{c}{PURE}  & 2.16 & 5.86 & 0.98\\
     ~&~ & \multicolumn{1}{c}{UBFC}  & 0.28 & 0.46 & 0.99\\
     \cmidrule(lr){2-6}
     ~&\multirow{2}{*}{PURE} & \multicolumn{1}{c}{PURE}  & 1.43 & 3.82 & 0.99\\
     ~&~ & \multicolumn{1}{c}{UBFC}  & 0.20 & 0.24 & 0.99\\
      \cmidrule(lr){2-6}
       ~&\multirow{2}{*}{UBFC} & \multicolumn{1}{c}{PURE}  & 0.40 & 0.92 & 0.99\\
     ~&~ & \multicolumn{1}{c}{UBFC}  & 0.17 & 0.21 & 0.99\\
     \bottomrule
     \end{tabular}}
     \vspace{-0.8em}
\end{table}

\begin{table}[t]\large
    \centering
    \caption{Semi-supervised HR estimation results of our method on VIPL-HR.} 
    \label{tab:semi}
    \resizebox{0.45\textwidth}{!} {
    \begin{tabular} {p{3cm} p{3cm} c c c }
     \toprule
      \multicolumn{1}{c}{Train Data} &  \multicolumn{1}{c}{Train Data} & \multicolumn{3}{c}{HR (bpm)}\\
      \cmidrule(lr){3-5}
     \multicolumn{1}{c}{w. Label} &  \multicolumn{1}{c}{w/o Label}& \multicolumn{1}{c}{MeanAE $\downarrow$}& \multicolumn{1}{c}{RMSE $\downarrow$} & \multicolumn{1}{c}{$r$ $\uparrow$}\\
     \midrule
     \multicolumn{1}{c}{10$\%$ $D_{train}$} &\multicolumn{1}{c}{/} & 9.40& 13.20 & 0.05\\
     \multicolumn{1}{c}{10$\%$ $D_{train}$} &\multicolumn{1}{c}{90$\%$ $D_{train}$} & 9.01& 12.69 & 0.35\\
     \multicolumn{1}{c}{20$\%$ $D_{train}$} &\multicolumn{1}{c}{/} & 9.67& 13.70 & 0.04\\
     \multicolumn{1}{c}{20$\%$ $D_{train}$} &\multicolumn{1}{c}{80$\%$ $D_{train}$} & 8.53& 12.19 & 0.35\\
     \multicolumn{1}{c}{50$\%$ $D_{train}$} &\multicolumn{1}{c}{/} & 8.08& 11.37 & 0.49\\
     \multicolumn{1}{c}{50$\%$ $D_{train}$} &\multicolumn{1}{c}{50$\%$ $D_{train}$}& 6.54& 9.90 & 0.63\\
     \bottomrule
     \end{tabular}
     }
     \vspace{-0.8em}
\end{table}

\textbf{HRV and RF estimation on UBFC-rPPG.}  We also conduct experiments for three types of physiological signals, i.e., HR, RF, and HRV measurement on UBFC-rPPG dataset. We compared our approach with three traditional methods, two supervised methods and two self-supervised methods, as shown in Table \ref{tab:HRV}. LF, HF, and LF/HF are three measures for HRV estimation. The proposed method gives more accurate estimation in terms of RF, and LF/HF than the SOTA method, which means that the proposed method could not only work on HR estimation task but also give a promising rPPG signal prediction for RF measurement and HRV analysis. 

\textbf{Transfer learning.}  We conduct a transfer learning task on three VIPL-HR, PURE, UBFC-rPPG datasets, and also ImageNet. The experiments were all conducted on the original STMap. From the Table \ref{tab:TransferL}, we can see that the MAE results of the proposed method, pre-training with rPPG domain dataset is significantly lower than pre-training with ImageNet dataset. In terms of the details, the results of pre-training with VIPL-HR, PURE or UBFC-rPPG, and finetuning with PURE indicate that larger datasets do not yield a better result in rPPG pre-training tasks. Conversely, pre-training with the smaller but higher quality UBFC-rPPG dataset obtains better performance than a larger VIPL-HR dataset. As shown in Fig. \ref{fig:dataset}, UBFC-rPPG is a dataset with bright lights and simple background. Compared to UBFC-rPPG, PURE has more complex backgrounds. More challenging, VIPL-HR includes complex lighting changes and large head movements, which means it can be difficult to learn physiological information from the facial area during pre-training. This also explains why the other two datasets performed well with the original STMap pretraining alone, while VIPL-HR needed improvement. To sum up, the proposed self-supervised pre-training method has higher requirements on the quality of the dataset than the dataset size.

\textbf{Semi-supervised learning.} As shown in Table \ref{tab:semi}, under a small amount of training data with label, adding self-supervised pre-training not only do help to lower the results of MeanAE, but benifits the encoder to learn the real physiological information of the facial video, as the Pearson's correlation coefficient improves a lot. When we increace the propotion of the labeled training data to 50$\%$, the Pearson's correlation coefficient of the row without pretraining then to be high. We suspect that MAE values that sometimes predict outcomes seem reasonable may be due to the network's accidental guesswork rather than a true learning of the nature of the data. The Pearson correlation coefficient with the increase of the proportion of pre-training data confirms that we have learned the inherent periodicity of physiological signals from the self-supervised pre-training.

\textbf{Linear probing.} Following \cite{LinearProbe}, we finetune the prediction head (a simple linear layer) with freezing the weights of self-supervised ViT encoder. As shown in Table \ref{tab:linear-probe}, our self-supervised ViT encoder pretrained with VIPL-HR shows a better performance. We draw three main conclusions: 1) By learning the differences between samples, comparative learning enables the model to extract the noise component that dominates the content rather than the physiological signal. 2) MAE method forces the model to learn the self-similarity within samples, which can better extract periodic physiological signals. 3) Pre-training on a small amount of STMap is better than on Imagenet, which indicates that there is a large domain gap between natural images and STMap.


\begin{table}[t]\Large
    \centering
    \caption{Linear-probing on VIPL-HR.} 
    \label{tab:linear-probe}
    \resizebox{0.45\textwidth}{!} {\begin{tabular}{p{3.5cm}  p{1.7cm} p{2.4cm} c c c} 
     \toprule
     \multirow{2}{*}{Method} &\multirow{2}{*}{Pretraining}&\multirow{2}{*}{linear-probing}&  \multicolumn{3}{c}{HR (bpm)}\\
     \cmidrule(lr){4-6}
     ~&~&~& \multicolumn{1}{c}{MeanAE $\downarrow$} & \multicolumn{1}{c}{RMSE $\downarrow$} & \multicolumn{1}{c}{$r$ $\uparrow$}  \\
     \midrule
     MoCo~\cite{2020MoCo}&\multicolumn{1}{c}{VIPL}& \multicolumn{1}{c}{VIPL} & 9.27 & 13.05 & 0.04  \\ 
     SIMSIAM~\cite{2021simsiam}&\multicolumn{1}{c}{VIPL}& \multicolumn{1}{c}{VIPL} & 8.43 & 11.73 & 0.14  \\ 
     BOYL~\cite{grill2020byol}&\multicolumn{1}{c}{VIPL}& \multicolumn{1}{c}{VIPL} & 8.98 & 12.43 & 0.08  \\ 
     SIMCLR~\cite{chen2020simclr}&\multicolumn{1}{c}{VIPL}& \multicolumn{1}{c}{VIPL} & 8.57 & 11.94 & 0.10  \\ 
     \midrule
    \multirow{2}{*}{\textbf{rPPG-MAE (Ours)}}&\multicolumn{1}{c}{ImageNet}& \multicolumn{1}{c}{VIPL} & 8.92 & 12.28 & 0.28  \\ 
    ~&\multicolumn{1}{c}{VIPL}& \multicolumn{1}{c}{VIPL} & \textbf{7.83} & \textbf{11.19} & \textbf{0.48}\\
     \bottomrule
     \end{tabular}}
\end{table}
\subsection{Cross-dataset Testing}
\label{sec:cross-dataset}
To evaluate the generalization of the methods, we perform cross-dataset testing on PURE and UBFC-rPPG following the protocol of~\cite{STMap2020}. First, we train our model on PURE, and test it on UBFC-rPPG. The results are shown in Table \ref{tab:cross-UBFC}, where results of all the compared methods are directly come from \cite{Dual-GAN2021}. Then, we also train our model as well as re-implement four methods (PhysNet~\cite{PhysNet}, MTTS-CAN~\cite{MTTS-CAN}, PhysFormer~\cite{PhysFormer}, and Instantaneous~\cite{Instantaneous}) on PURE, and test it on UBFC-rPPG. The results are shown in Table \ref{tab:cross-PURE}. It is clear that the proposed method outperforms other methods by a large margin on cross-testing from UBFC-rPPG to PURE, indicating the strong generalization ability of rPPG-MAE.

  \begin{table}[t]\small
    \centering
    \caption{Cross-dataset HR estimation (training on PURE and
    testing on UBFC-rPPG).} 
    \label{tab:cross-UBFC}
    \resizebox{0.4\textwidth}{!} {\begin{tabular} {p{3.2cm}p{0.2cm}p{0.2cm}} 
     \toprule
      \multirow{2}{*}{Method}& \multicolumn{1}{c}{MeanAE $\downarrow$}& \multicolumn{1}{c}{RMSE $\downarrow$} \\
     ~&(bpm)&(bpm)\\
     \midrule
    GREEN~\cite{GREEN2008} & 8.29&  15.82\\
     ICA~\cite{ICA2011} & 4.39& 11.60\\
     POS~\cite{POS2014} & 3.52& 8.38\\
     CHROM~\cite{CHROM2013} & 3.10& 6.84\\
     PulseGAN~\cite{PulseGAN} & 2.09& 4.42\\
     Dual-GAN~\cite{Dual-GAN2021} & \textbf{0.74}& \textbf{1.02}\\
     \textbf{rPPG-MAE (Ours)} & \underline{1.28}& \underline{2.75}\\
     \bottomrule
     \end{tabular}}
\end{table}
\begin{table}[t]\small
    \centering
    \caption{Cross-dataset HR estimation (training on UBFC-rPPG and
    testing on PURE).} 
    \label{tab:cross-PURE}
    \resizebox{0.4\textwidth}{!} {\begin{tabular} {p{3.2cm}p{0.2cm}p{0.2cm}} 
     \toprule
      \multirow{2}{*}{Method}& \multicolumn{1}{c}{MeanAE $\downarrow$}& \multicolumn{1}{c}{RMSE $\downarrow$} \\
     ~&(bpm)&(bpm)\\
     \midrule
   PhysNet~\cite{PhysNet} & 31.45& 39.25 \\
     MTTS-CAN~\cite{MTTS-CAN} & \underline{16.77}& \underline{31.28}\\
    PhysFormer~\cite{PhysFormer} & 23.63& 30.70\\
     Instantaneous~\cite{Instantaneous} & 33.84& 35.68\\
     \textbf{rPPG-MAE (Ours)} & \textbf{13.55}& \textbf{20.27}\\
     \bottomrule
     \end{tabular}}
\end{table}





\subsection{Ablation Studies}
\label{sec:Ablation Studies}
We provide the results of ablation studies for HR estimation on the VIPL-HR~\cite{VIPL} datatset, and the protocol of this part is the same as Section \ref{sec:Details} fine-tuning.

\begin{figure}
\centering
\includegraphics[scale=0.55]{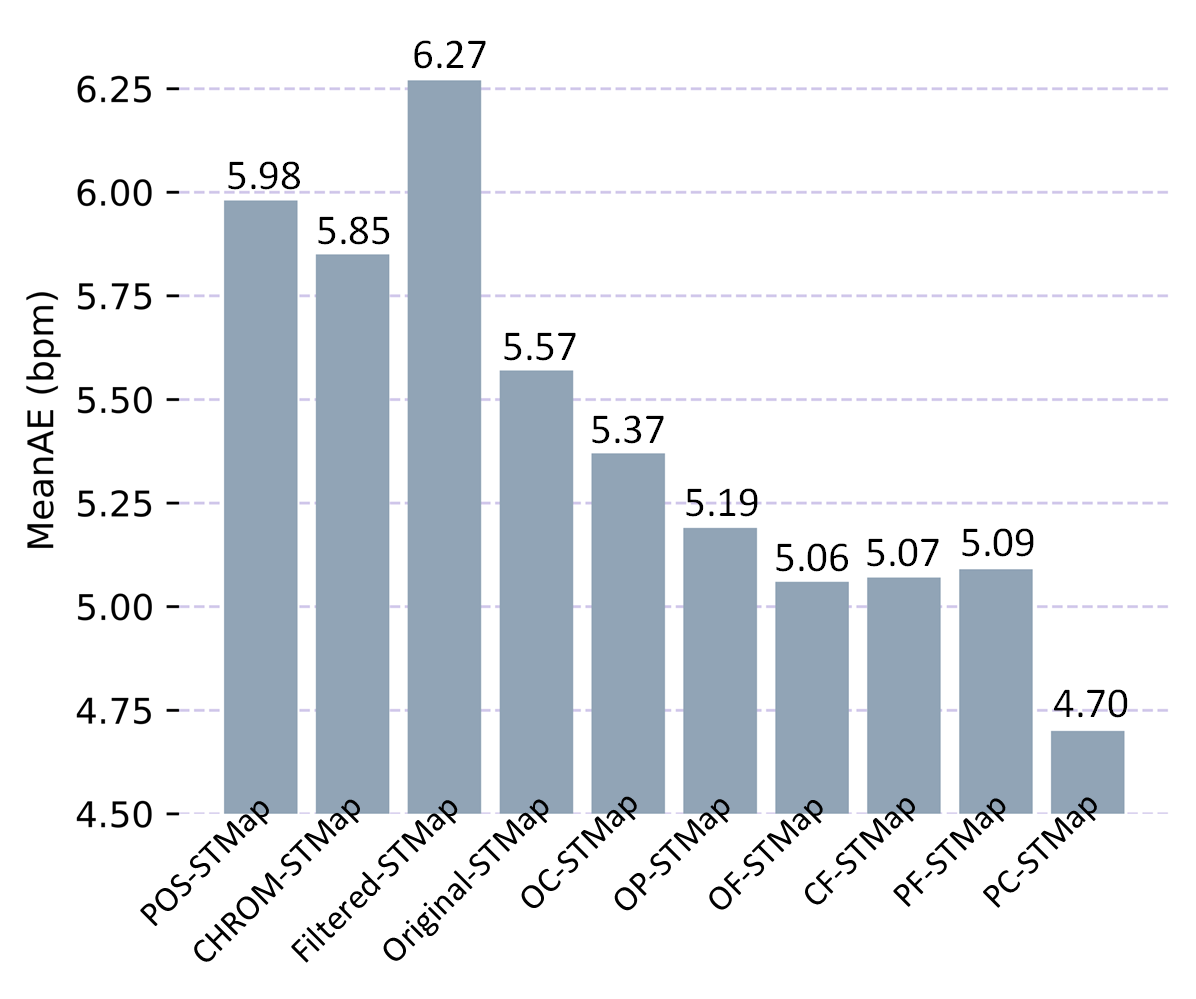}
\vspace{-0.8em}
  \caption{\small{\textbf{Ablation on map type.} The y-axes are mean-absolute-error values fine-tuned on different STMap with 100 epochs pre-trained.
  }
  }
\label{fig:results_of_map_type}
\vspace{-1.0em}
\end{figure} 

\textbf{Impact of STMap types for reconstruction.} As shown in Fig. \ref{fig:results_of_map_type}, from the four STMaps of one single type, we can see that compared with the other three STmaps (POS-STMap, CHROM-STMap, Filtered-STMap), the original STMap is more conducive to the encoder to learn the periodic information of rPPG signal, as the MeanAE is the lowest among the four. We hypothesize that mask autoencoder requires a greater degree of preservation of the structure and information of the original image, which needs to preserve the structure and information of the original image to a greater extent. Interestingly, when we combined the two better STMaps, its (OC-STMap) results did not perform the best among the remaining six hybrid STMaps. Surprisingly, the Filtered-STMap combined with the Original-STMap show better performance than the other two combinations (OC-STMap and OP-STMap).This can be interpreted as Filtered-STMap may lose some necessary physiological information when filtering out the noise. Therefore, the encoder cannot learn the physiological information well when it is reconstructed separately. However, when it supplements with the Original-STMap, its function of filtering the noise is revealed. The best performance is achieved by the PC-STMap, due to the synergistic effect: 1) the improved motion robustness brought by the POS and CHROM; 2) the combination of different color space reduces sensitivity to changes in illumination.

\begin{figure}
\centering
\includegraphics[scale=0.25]{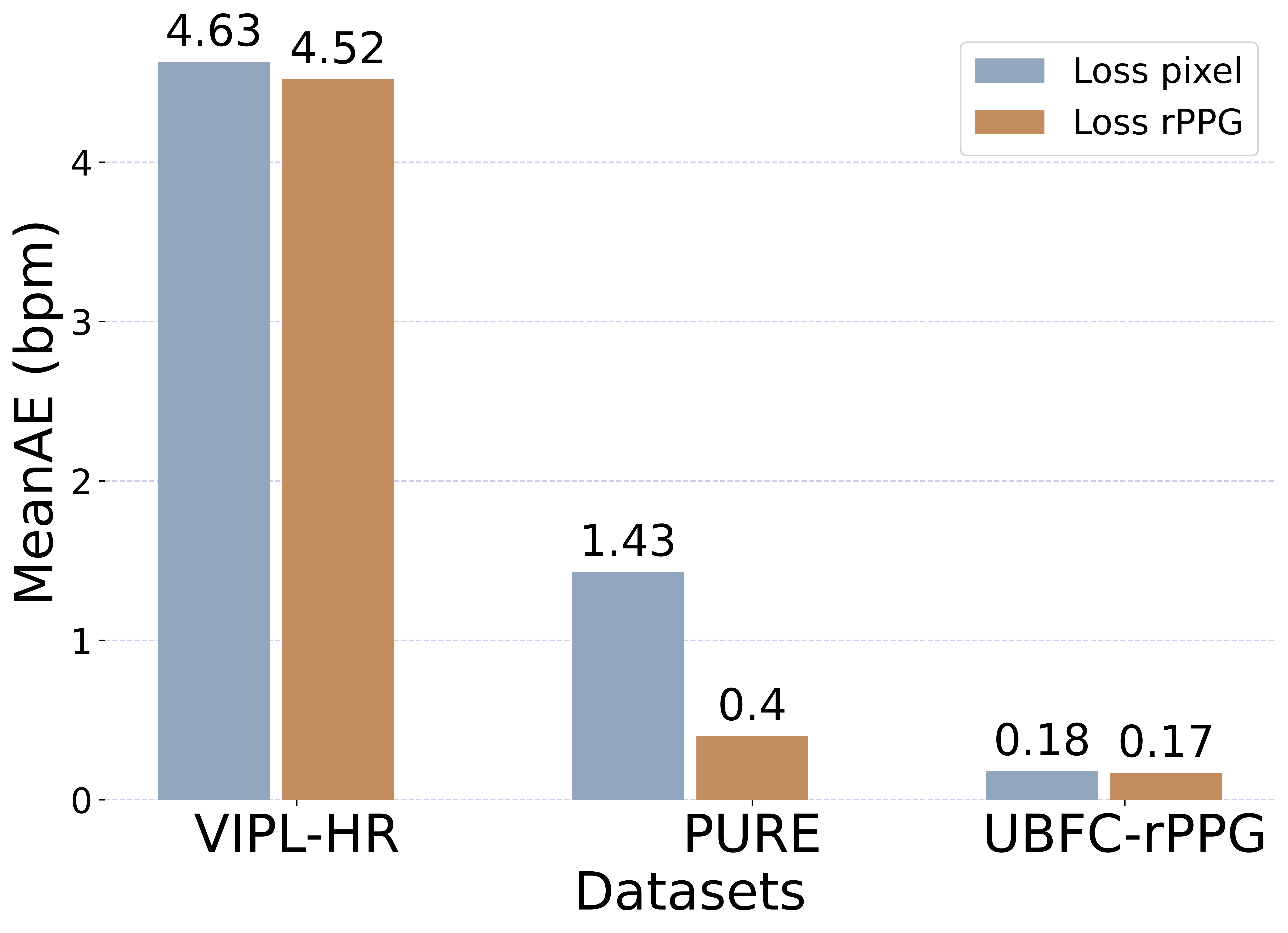}
\vspace{-0.8em}
  \caption{\small{\textbf{Ablation on loss type.} The y-axes
are mean-absolute-error values fine-tuned on different dataset.
  }
  }
\label{fig:losstype}
\vspace{-1.0em}
\end{figure} 
\textbf{Impact of pre-training loss type.} We conduct experiments with two reconstruction loss functions which introduced in Section \ref{sec:rPPG Finetuning}. As shown in Fig. \ref{fig:losstype}, experiments indicate that both loss functions are effective, while the proposed rPPG loss function outperforms original pixel-to-pixel loss. Especially on PURE, the improvement is remarkable.

\begin{figure}
    \centering
    \includegraphics[scale=0.35]{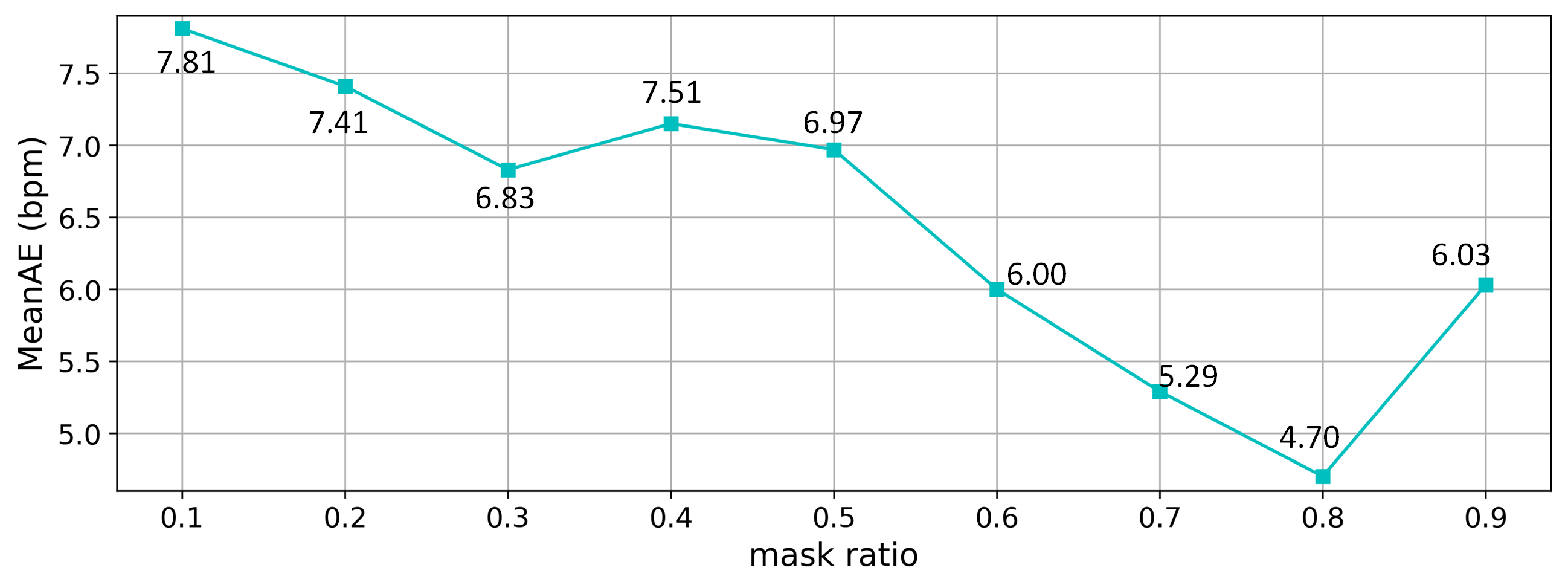}
      \caption{\small{Results with different masking ratio. The y-axes are mean-absolute-error values fine-tuned on VIPL-HR with 100 epochs pre-trained.
      }
      }
    \label{fig:maskratio}
\end{figure} 

\textbf{Impact of masking ratio.} To see the experimental results with different masking ratio, the ratio setting in pre-training stage is adjusted. As illustrated in Fig. \ref{fig:maskratio}, with the increase of mask ratio, the experimental score increased first and then decreased, and the best score is achieved with the mask ratio $80\%$. It verified that our rPPG-MAE learn the physiological information in STMap with a very small proportion of patches.

\begin{figure}
\centering
\includegraphics[scale=0.5]{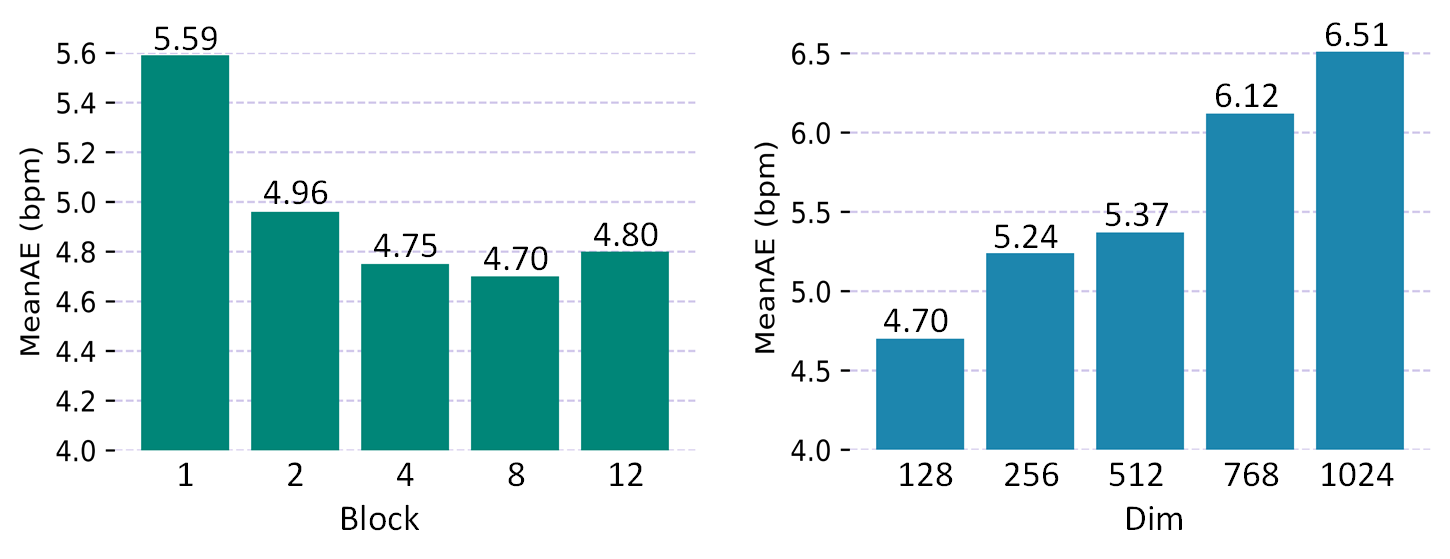}
  \caption{\small{Ablation on decoder.The y-axes are mean-absolute-error values fine-tuned on VIPL-HR with 100 epochs pre-trained. The default set is 8 blocks andd 128 dim.
  }
  }
\label{fig:block}
\vspace{-0.4em}
\end{figure} 

\textbf{Decoder design.} Our ViT decoder can be flexible designed, we adjust the number of Transformer blocks and the number of channels, respectively. As shown in Fig. \ref{fig:block}, a deep and narrow decoder works better in our rPPG task, which can be interpreted as a lightweight decoder benefits the the network to train the encoder rather than the decoder. The default decoder of rPPG-MAE has 8 blocks and a width of 128-dimension.

\begin{figure}
\centering
\includegraphics[scale=0.35]{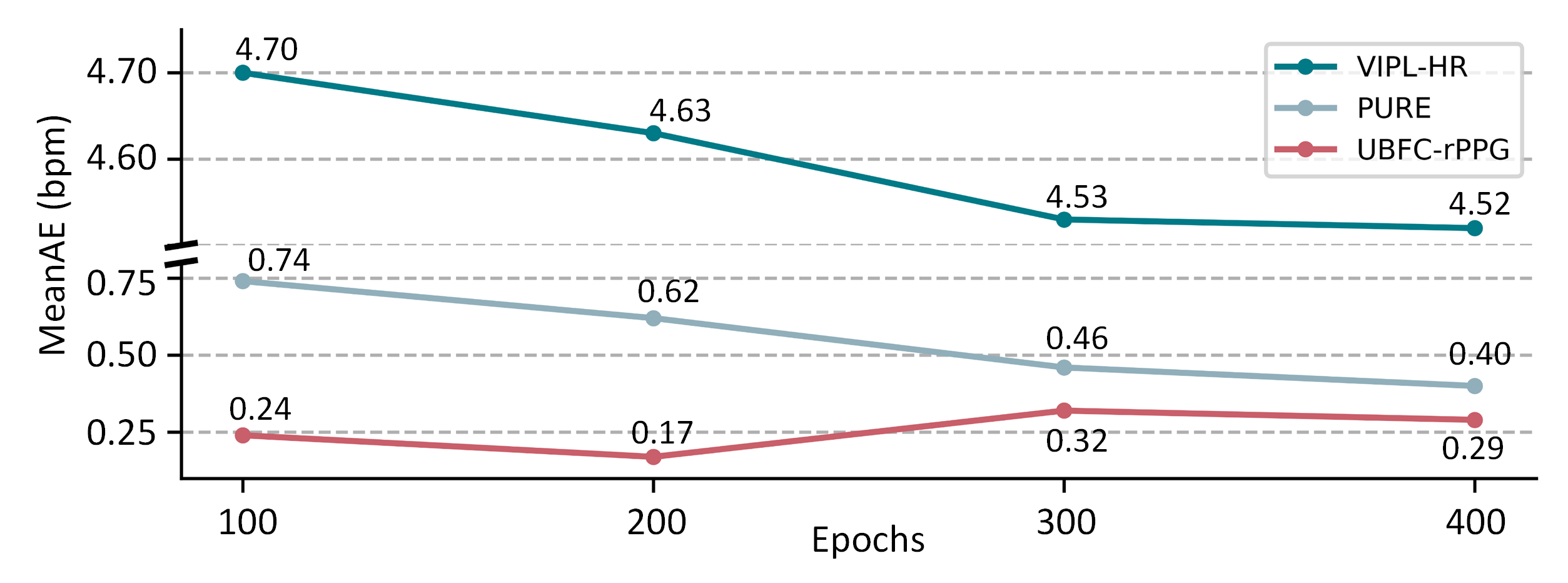}
\vspace{-0.8em}
  \caption{\small{\textbf{Training schedule.} The y-axes
are mean-absolute-error values fine-tuned on VIPL-HR.
  }
  }
\label{fig:trainingschrdule}
\vspace{-1.0em}
\end{figure} 

\textbf{Training schedule.} So far, our ablations are based on 100-epoch pre-training on VIPL-HR dataset. Fig. \ref{fig:trainingschrdule} shows the influence of the training schedule length. As train schedule length gets longer, the MeanAE on PURE and VIPL-HR keeps doing down and the performances on UBFC-rPPG fluctuate within a reasonable range. In fact, when the training schedule reached 100 epochs, the model achieved good performance on the three datasets. After that, adding the epoch of pre-training seems benefit little. There are two points that should be taken into consideration: 1) Compared to ImageNet, the size of dataset in rPPG is
much more smmaller. That's to say, a relatively pre-training deuration is enough. 2) Longer pre-training time means not only the learning of rPPG signals, but also finer image detail and noise capture, which is a trade-off.
 
\begin{figure*}
\centering
\includegraphics[scale=0.85]{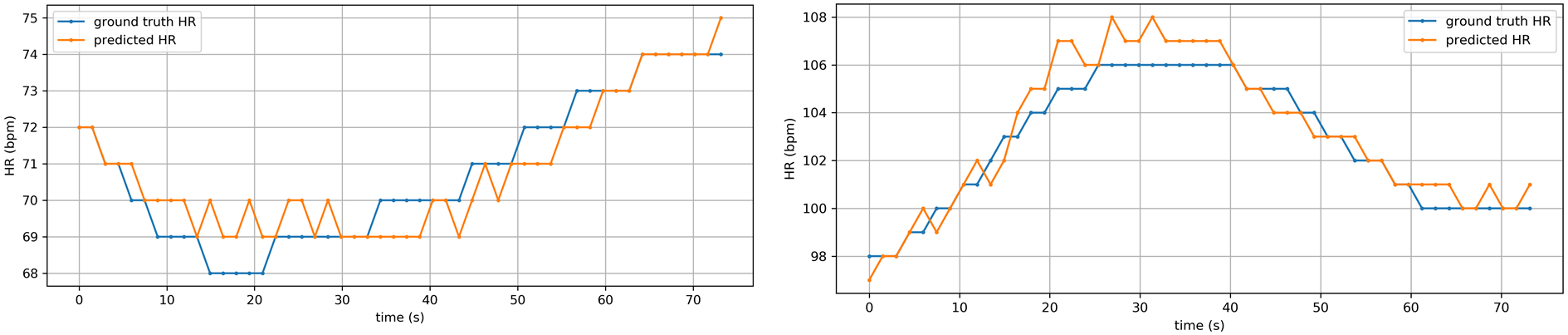}
  \caption{\small{Visualization of predicted rPPG signals on UBFC-rPPG and PURE.
  }
  }
\label{fig:HR}
\vspace{-0.4em}
\end{figure*} 

\begin{figure*}
\centering
\includegraphics[scale=0.9]{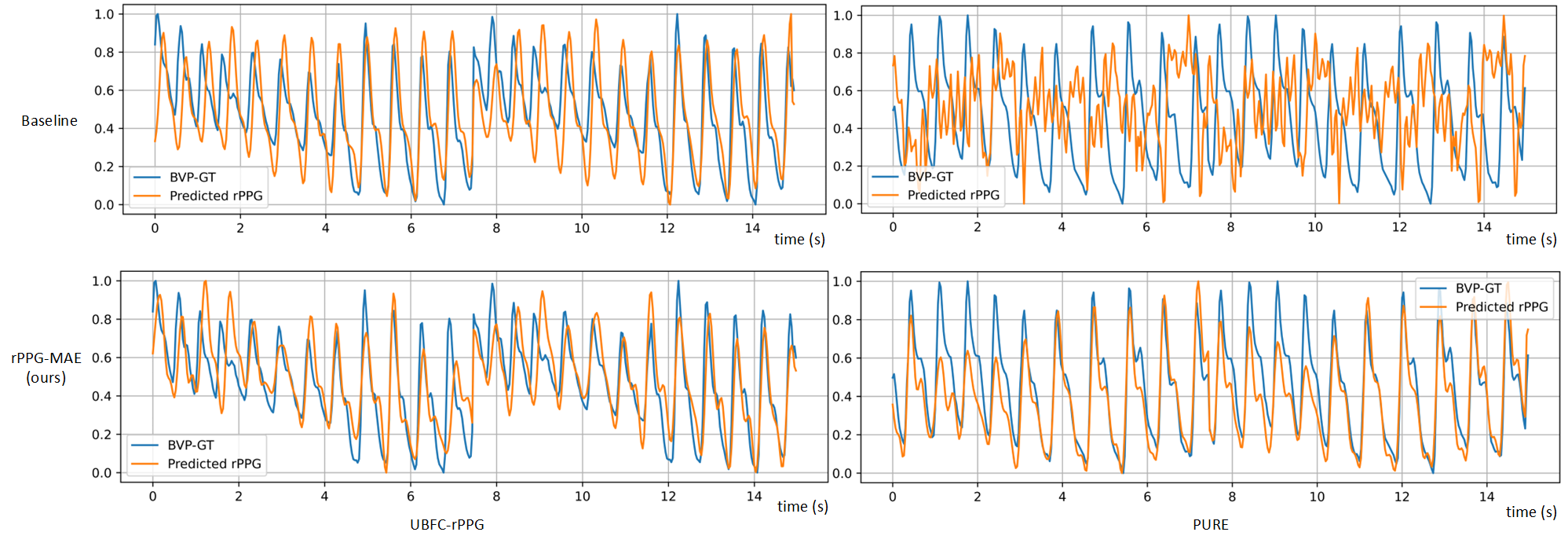}
  \caption{\small{Visualization of predicted rPPG signals on UBFC-rPPG and PURE.
  }
  }
\label{fig:Wave}
\vspace{-0.4em}
\end{figure*} 

 \begin{figure}
\centering
\includegraphics[scale=0.62]{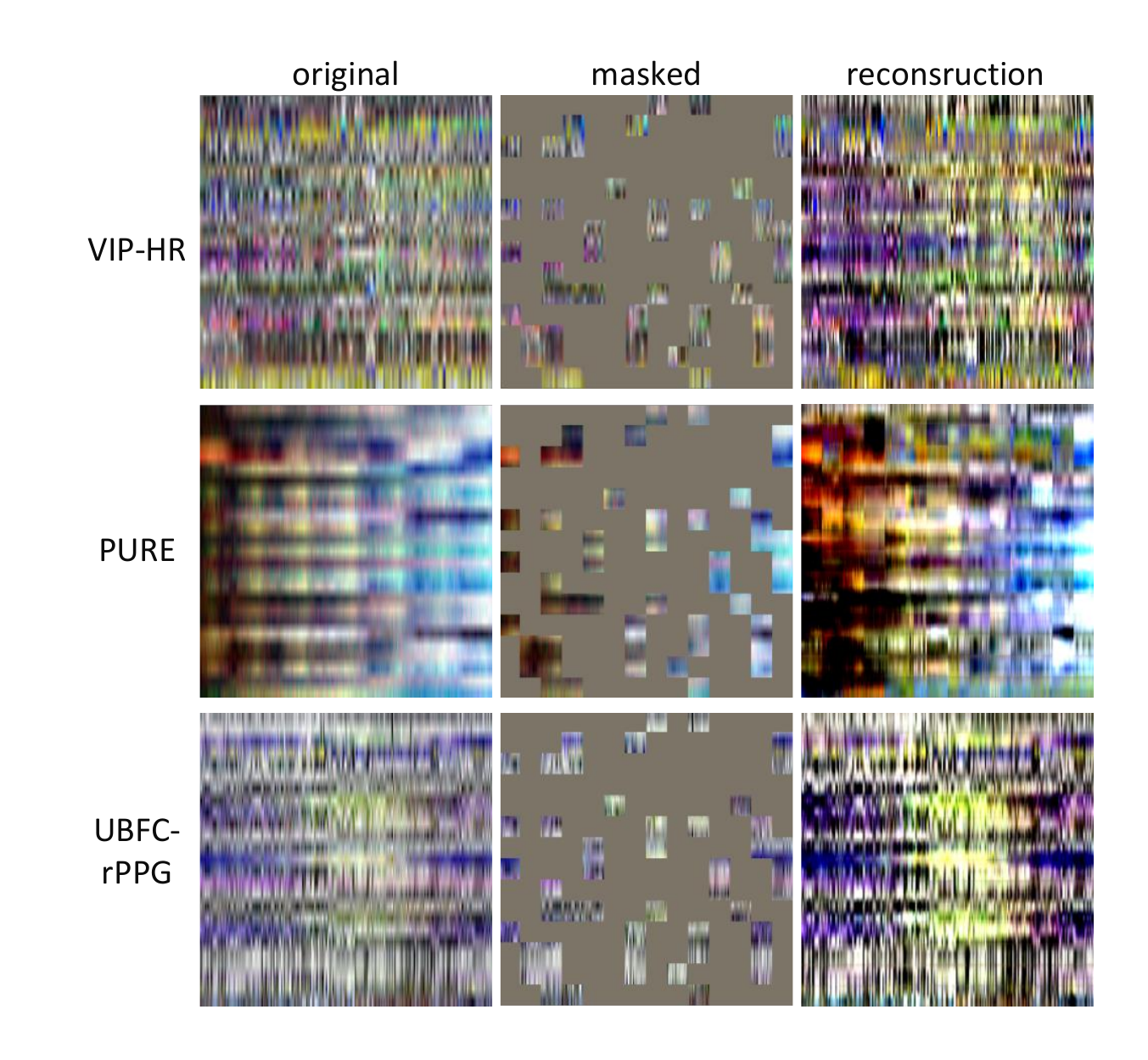}
  \caption{\small{Visualization of reconstructive STMaps. Each row of images comes from the same dataset. The three columns are the original STMap, masked STMap, and reconstructive STMap.
  }
  }
\label{fig:reconstruction}
\vspace{-0.4em}
\end{figure}


\subsection{Visualization and Discussion}
 \label{sec:Analysis}  
The examples in Fig. \ref{fig:HR} show that the predicted heart rate was almost identical to the ground truth. We also visualize the results of rPPG prediction in Fig. \ref{fig:Wave}, respectively. From the prediction results, the rPPG signal predicted by rPPG-MAE can better fit the ground-truth of BVP signal. As shown in Fig. \ref{fig:Wave}, the predicted rPPG signal by rPPG-MAE is much more closer to the ground truth BVP signals in both frequency and amplitude than the baseline on PURE. On a simper UBFC-rPPG dataset, the Baseline has performed well while rPPG-MAE is clearly better at fitting the signal amplitude. In Fig. \ref{fig:reconstruction}, we visualize the reconstructive STMaps from the ViT decoder with the proposed rPPG loss function pre-trained. It can be seen that the reconstructed STMap has a similar color texture to the original STMap, which indicates that periodicity of the intrinsic rPPG features has been learned by the network.

\vspace{-0.8em}
\section{Conclusion}
\label{sec:conclusion}
In this paper, we propose a self-supervised mask-autoencoder architecture, namely rPPG-MAE, for realize a efficient remote physiological measurement. With PC-STMap as input and the proposed rPPG loss function, rPPG-MAE is able to achieve superior performance on the challenging VIPL-HR dataset. We discuss the the depth and width of decoder, map types, pre-training schedule, and mask ratio of rPPG-MAE and find the most suitable architecture design. The structure of rPPG-MAE still has room for improvement. Future directions include: 1) Designing rPPG-MAE variants. Since the architecture of rPPG-MAE is similar to the image classification field, a more suitable MAE-based network for remote physiological measurement should be designed. 2) Adding data augmentation. Proper data augmentation is much helpful for self-supervised learning.

\vspace{-0.8em}

\ifCLASSOPTIONcaptionsoff
  \newpage
\fi

\bibliographystyle{IEEEtran}
\bibliography{IEEEabrv,reference}



\end{CJK}
\end{document}